\documentclass[1p]{elsarticle}
\usepackage{amsmath,amssymb,bm,graphicx,color}
\usepackage{lineno}
\usepackage[colorlinks, linkcolor=red, anchorcolor=blue, citecolor=green]{hyperref}
\usepackage[titletoc]{appendix}
\begin{document}
	
	\begin{frontmatter}
		
		\title{Understanding and Comparing Scalable Gaussian Process Regression for Big Data}
		
\author[mymainaddress]{Haitao Liu\corref{mycorrespondingauthor}}
\cortext[mycorrespondingauthor]{Corresponding author}
\ead{htliu@ntu.edu.sg}

\author[mysecondaryaddress]{Jianfei Cai}
\ead{ASJFCai@ntu.edu.sg}

\author[mysecondaryaddress,mythirdaddress]{Yew-Soon Ong}
\ead{ASYSOng@ntu.edu.sg}

\author[myforthaddress]{Yi Wang}
\ead{Yi.Wang4@Rolls-Royce.com}

\address[mymainaddress]{Rolls-Royce@NTU Corporate Lab, Nanyang Technological University, Singapore 637460}
\address[mysecondaryaddress]{School of Computer Science and Engineering, Nanyang Technological University, Singapore 639798}
\address[mythirdaddress]{Data Science and Artificial Intelligence Research Center, Nanyang Technological University, Singapore 639798}
\address[myforthaddress]{Central Technology Group, Rolls-Royce Singapore, 1 Seletar Aerospace Crescent, Singapore 797565}

\begin{abstract}
As a non-parametric Bayesian model which produces informative predictive distribution, Gaussian process (GP) has been widely used in various fields, like regression, classification and optimization. The cubic complexity of standard GP however leads to poor scalability, which poses challenges in the era of big data. Hence, various scalable GPs have been developed in the literature in order to improve the scalability while retaining desirable prediction accuracy. This paper devotes to investigating the methodological characteristics and performance of representative global and local scalable GPs including sparse approximations and local aggregations from four main perspectives: scalability, capability, controllability and robustness. The numerical experiments on two toy examples and five real-world datasets with up to 250K points offer the following findings. In terms of scalability, most of the scalable GPs own a time complexity that is linear to the training size. In terms of capability, the sparse approximations capture the long-term spatial correlations, the local aggregations capture the local patterns but suffer from over-fitting in some scenarios. In terms of controllability, we could improve the performance of sparse approximations by simply increasing the inducing size. But this is not the case for local aggregations. In terms of robustness, local aggregations are robust to various initializations of hyperparameters due to the local attention mechanism. Finally, we highlight that the proper hybrid of global and local scalable GPs may be a promising way to improve both the model capability and scalability for big data.
\end{abstract}


\end{frontmatter}

%\linenumbers

\section{Introduction}
Surrogate-assisted modeling and optimization have been extensively deployed to facilitate modern aeroengine design~\cite{duchaine2009computational, liu2014modeling, amrit2016efficient, wagle2017forward} due to the representational capability of complex features. Among current surrogates (also known as machine learning models), as a non-parametric Bayesian model, Gaussian process (GP)~\cite{rasmussen2006gaussian} (also known as Kriging or emulator), has gained popularity.

Given $n$ training data points $\bm{X} = \{\bm{x}_i \in R^d\}_{i=1}^n$ and the relevant observations $\bm{y} = \{y_i \in R\}_{i=1}^n$, GP intends to infer the latent function $f: R^d \mapsto R$, which is drawn from a Gaussian process, to describe the data pattern. Compared to other popular machine learning models, e.g., random forest~\cite{liaw2002classification} and deep neural networks~\cite{lecun2015deep}, the Bayesian perspective allows GP to provide not only the predictions at unseen points but also the uncertainties about the predictions.\footnote{It has been pointed out that GP is equivalent to a shallow but infinitely wide neural network with Gaussian weights~\cite{neal2012bayesian}.} The informative predictive distribution enables GP to be deployed in various scenarios, e.g., active learning~\cite{liu2018survey}, time-series forecast~\cite{foreman2017fast} and Bayesian optimization~\cite{snoek2015scalable}. However, an inherent disadvantage of GP is that it scales poorly with the training size $n$, since as a kernel method it handles the data in a one-shot fashion. The standard implementation of GP needs to invert an $n \times n$ covariance matrix, resulting in $\mathcal{O}(n^2)$ complexity in storage and $\mathcal{O}(n^3)$ complexity in time. Hence, the full GP becomes intractable when the training size is greater than $\mathcal{O}(10^4)$~\cite{deisenroth2015distributed}. 

In the era of big data,\footnote{Big data here mainly refers to the datasets with extremely large sample size $n$ such that they are expensive to be stored and analyzed.} numerous data brings vast quantity of information to enhance the analysis, learning and exploration in the machine learning community. Hence, there is a great demand for improving the scalability of standard GP while retaining desirable prediction accuracy.

In current literature, various scalable GPs have been developed for handing big data in different ways. They usually can be classified into two main categories:
\begin{itemize}
	\item \textit{Global approximations} that summarize the whole training data using a small subset. The simplest way is the subset-of-data (SoD) which trains GP on a random subset of the training data $\mathcal{D}=\{\bm{X},\bm{y}\}$~\cite{chalupka2013framework}. The performance of SoD however is limited due to the ignorance of the remaining data. The most commonly used global approximations are the so-called sparse approximations~\cite{quinonero2005unifying, snelson2006sparse, titsias2009variational, hensman2013gaussian}, which employ a set of global inducing variables to approximate the joint prior or the interested posterior. By using $m$ inducing points, sparse approximations reduce the time complexity of full GP to $\mathcal{O}(nm^2)$~\cite{snelson2006sparse, titsias2009variational} and more remarkably, $\mathcal{O}(m^3)$~\cite{hensman2013gaussian}. The advantages of sparse approximations are that they are usually derived in a unifying and elegant Bayesian framework, thus yielding a complete probabilistic model. The limitation however is that the modeling performance is limited by the small set of global inducing points, i.e., it is difficult for them to capture the quick-varying features, especially in high dimensions~\cite{bui2014tree}.
	\item \textit{Local approximations} that employ the idea of divide-and-conquer (D\&C) to distribute the whole training process over multiple local GP experts. By partitioning the data $\mathcal{D}$ into $M$ local subsets $\{\mathcal{D}_i=\{\bm{X}_i, \bm{y}_i\} \}_{i=1}^M$, each of which is assumed to have the same training size $m_0 = n/M$, local approximations yield the time complexity of $\mathcal{O}(nm_0^2)$. The benefits brought by the D\&C idea are that (i) it enables capturing local patterns; and (ii) it scales local approximations up to arbitrary dataset due to the straightforward parallel/distributed structure.
	The simplest local approximation is to training individual GP experts on multiple local subsets, which however suffers from discontinuous predictions and local over-fitting. Hence, local aggregations have been presented to smooth the predictions by aggregating the predictions from multiple local experts~\cite{deisenroth2015distributed, hinton2002training, tresp2000bayesian, cao2014generalized, liu2018generalized}. The limitation of local aggregations however is that they cannot provide a complete probabilistic model, which results in the so-called Kolmogorov inconsistency~\cite{samo2016string}. 
\end{itemize}

This paper intends to comprehensively investigate the characteristics and performance of representative scalable GPs on real-world large-scale datasets. As for the global approximations, we mainly introduce and compare the sparse approximations. Particularly, the sparse approximations are classified into two categories: the \textit{prior approximations}~\cite{snelson2006sparse, smola2001sparse, seeger2003fast} which approximate the prior but perform exact inference, and the \textit{posterior approximations}~\cite{titsias2009variational, hensman2013gaussian, dezfouli2015scalable} which retain the exact prior but perform approximate inference. We do not investigate some particular sparse approximations, e.g., the structured kernel interpolation~\cite{wilson2015kernel} which exploits the inducing set with Kronecker structure, since the inducing size increases exponentially with the dimensionality $d$, and the training quickly becomes intractable when $d>5$. 

As for the local approximations, we introduce and compare the pure local GPs and the local aggregations using product-of-experts and Bayesian committe machine~\cite{deisenroth2015distributed, hinton2002training,tresp2000bayesian,cao2014generalized, liu2018generalized}. We here do not investigate the mixture-of-experts~\cite{rasmussen2002infinite,yuksel2012twenty} since it is mainly designed for capturing non-stationary features and suffers from intractable inference.

The comparative study introduces and investigates the characteristics of representative scalable GPs from four main perspectives:
\begin{itemize}
	\item \textit{Scalability} that indicates the time complexity of scalable GPs to handle big data;
	\item \textit{Capability} that means the representational ability of scalable GPs to handle various tasks;
	\item \textit{Controllability} that indicates whether we can easily improve the performance of scalable GPs by tuning model parameters, e.g., the inducing size or the number of experts;
	\item \textit{Robustness} means the sensitivity of scalable GPs to various initializations of hyperparameters.
\end{itemize}

The remainder of the paper is organized as follows. Section~\ref{sec_GP} briefly introduces the standard GP. Thereby, Sections~\ref{sec_global} and~\ref{sec_local} respectively introduce the global and local approximations. Thereafter, Section~\ref{sec_exps} comprehensively investigates the characteristics and performance of these scalable GPs on several toy examples and five real-world large-scale datasets with up to 250K data points. Finally, Section~\ref{sec_cons} provides concluding remarks.

\section{Gaussian processes regression}
\label{sec_GP}
\begin{figure}[hbt!] 
	\centering
	\includegraphics[width=0.8\textwidth]{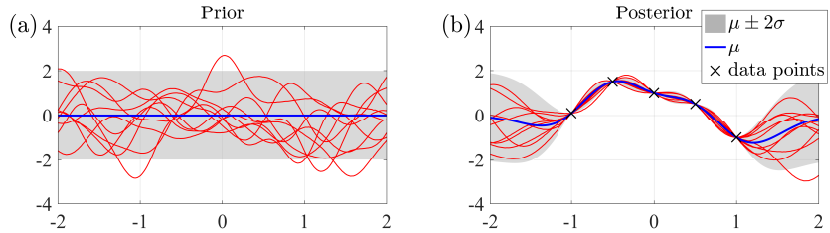}
	\caption{Samples (red curves) drawn from (a) the zero-mean GP prior and (b) the posterior after observing four data points.}
	\label{Fig_GP_prior_posterior} 
\end{figure}

As a non-parametric Bayesian model, the Gaussian process places a GP prior over the latent function $f: R^d \mapsto R$ as
\begin{equation}
f(\bm{x}) = \mathcal{GP}(m(\bm{x}), k(\bm{x}, \bm{x}')),
\end{equation}
where $m(.)$ is the mean function which is usually taken as zero without loss of generality, and $k(.,.)$ is the kernel function which controls the model smoothness. Popularly, we use the well-known squared exponential (SE) kernel function with automatic relevance determination (ARD) as
\begin{equation} \label{eq_SE_Kernel}
k(\bm{x},\bm{x}')= \sigma^2_{f} \exp \left(- \frac{1}{2} \sum_{i=1}^d \frac{(x_i-x'_i)^2}{l_i^2} \right),
\end{equation}
where $\sigma^2_{f}$ is the output signal, and $l_i$ is the input length-scale along the $i$th dimension. Fig.~\ref{Fig_GP_prior_posterior}(a) illustrates several samples drawn from the zero-mean GP prior.

Let us consider a regression task $y(\bm{x}) = f(\bm{x}) + \epsilon$ where $\epsilon \sim \mathcal{N}(0, \sigma^2_{\epsilon})$ is the \textit{iid} noise, the likelihood writes $p(y|f) = \mathcal{N}(y|f, \sigma^2_{\epsilon})$. Typically, the GP can be trained by finding the optimal hyperparameters $\bm{\theta} = \{\sigma^2_f, l_1, \cdots, l_d, \sigma^2_{\epsilon} \}$ through maximizing the log marginal likelihood (model evidence) as
\begin{equation} \label{eq_logp(y)}
\begin{aligned}
\log p(\bm{y}) = -\frac{n}{2} \log(2 \pi) - \frac{1}{2} \log |\bm{K}_{nn} + \sigma^2_{\epsilon} \bm{I}| - \frac{1}{2} \bm{y}^{\mathsf{T}} (\bm{K}_{nn} + \sigma^2_{\epsilon} \bm{I})^{-1} \bm{y},
\end{aligned}
\end{equation}
where $\bm{K}_{nn} = k(\bm{X}, \bm{X})$ is the $n \times n$ kernel matrix. Conditioned on the training data and the inferred hyperparameters, we derive the predictive (posterior) distribution $p(f_*|\mathcal{D}, \bm{x}_*)$ at a test point $\bm{x}_*$ with the mean and variance respectively given as
\begin{subequations}
	\begin{align}
	\mu_* =& k_{*n} [\bm{K}_{nn} + \sigma^2_{\epsilon} \bm{I}]^{-1} \bm{y}, \\
	\sigma^2_* =& k_{**} - \bm{k}_{*n} [\bm{K}_{nn} + \sigma^2_{\epsilon} \bm{I}]^{-1} \bm{k}_{*n}^{\mathsf{T}},
	\end{align}
\end{subequations}
where $k_{**} = k(\bm{x}_*, \bm{x}_*)$ and $\bm{k}_{*n} = k(\bm{x}_*, \bm{X})$. Hence, the predictive distribution of $y_*$ has $p(y_*|\mathcal{D}, \bm{x}_*) = \mathcal{N}(y_*|\mu_*, \sigma^2_* + \sigma^2_{\epsilon})$. Fig.~\ref{Fig_GP_prior_posterior}(b) depicts several samples drawn from the posterior after observing four data points.

It is found in~\eqref{eq_logp(y)} that the computational bottleneck in the full GP inference is the inversion $(\bm{K}_{nn} + \sigma^2_{\epsilon} \bm{I})^{-1}$ and the determinant $|\bm{K}_{nn} + \sigma^2_{\epsilon} \bm{I}|$, both of which require the time complexity of $\mathcal{O}(n^3)$ through standard Cholesky decomposition.

The cubic complexity poses urgent demand of improving the scalability of standard GP for handling big data. In what follows, we will introduce representative scalable GPs including 
\begin{itemize}
	\item \textit{global approximations}, particularly the sparse approximations, which distillate the training data through $m$ inducing points; and
	\item \textit{local approximations}, particularly the local aggregations, which follow the idea of D\&C to aggregate multiple GP experts for boosting predictions.
\end{itemize}

\section{Global approximations}
\label{sec_global}
The simplest global approximation is the subset-of-data (SoD) which trains GP on a random subset of the training data. Given $m$ ($m \ll n$) subset points, the time complexity of SoD is substantially reduced to $\mathcal{O}(m^3)$ in comparison to the full GP. The SoD however may produce over-confident predictions by ignoring the remaining data.

To date, the popular global approximations are sparse approximations~\cite{quinonero2005unifying} motivated by the Nystr\"{o}m approximation~\cite{williams2001using}. Sparse approximations employ an inducing set $\bm{X}_m$ which includes $m$ points  to summarize the whole training data, thus reducing the time complexity to $\mathcal{O}(nm^2)$.\footnote{The inducing points in $\bm{X}_m$ are usually regarded as hyperparameters to be inferred.} The latent inducing variables $\bm{f}_m$ akin to $\bm{f}$ follow the same GP prior $p(\bm{f}_m) = \mathcal{N}(\bm{f}_m|\bm{0}, \bm{K}_{mm})$. 

In what follows, current sparse approximations are classified into two main categories: the \textit{prior approximations} which approximate the prior but perform exact inference, and the \textit{posterior approximations} which retain the exact prior but perform approximate inference.

\subsection{Prior approximations}
Given the independence assumption $\bm{f} \perp f_*| \bm{f}_m$, the joint prior $p(\bm{f}, f_*)$ can be derived by marginalizing out $\bm{f}_m$ as
\begin{equation} \label{eq_p(f,f*)}
p(\bm{f}, f_*) = \int p(\bm{f}| \bm{f}_m) p(f_*|\bm{f}_m) p(\bm{f}_m) d\bm{f}_m.
\end{equation}
Given the Nystr\"{o}m notation $\bm{Q}_{ab} = \bm{K}_{am} \bm{K}_{mm}^{-1} \bm{K}_{mb}$, the training and test conditionals in~\eqref{eq_p(f,f*)} are respectively expressed as
\begin{subequations}
	\begin{align}
	p(\bm{f} | \bm{f}_m) =& \mathcal{N} (\bm{f}|\bm{K}_{nm} \bm{K}_{mm}^{-1} \bm{f}_m, \bm{K}_{nn} - \bm{Q}_{nn}), \label{eq_training_cond} \\
	p(f_* | \bm{f}_m) =& \mathcal{N} (f_*|\bm{k}_{*m} \bm{K}_{mm}^{-1} \bm{f}_m, k_{**} - Q_{**}). \label{eq_test_cond} 
	\end{align}
\end{subequations}
Now we clearly see that $\bm{f}_m$ is called inducing variables since it induces the dependence between the independent $\bm{f}_m$ and $f_*$.  

To achieve computational gains, prior approximations further modify the joint prior as 
\begin{equation} \label{eq_modify_prior}
p(\bm{f}, f_*) \approx q(\bm{f}, f_*) = \int q(\bm{f}| \bm{f}_m) q(f_*|\bm{f}_m) p(\bm{f}_m) d\bm{f}_m.
\end{equation}
Compared to~\eqref{eq_p(f,f*)}, the training and test conditionals are respectively approximated in~\eqref{eq_modify_prior} through replacing the co-variance matrices as
\begin{subequations}
	\begin{align}
	q(\bm{f} | \bm{f}_m) =& \mathcal{N} (\bm{f}|\bm{K}_{nm} \bm{K}_{mm}^{-1} \bm{f}_m, \tilde{\bm{Q}}_{nn}), \\
	q(f_* | \bm{f}_m) =& \mathcal{N} (f_*|\bm{k}_{*m} \bm{K}_{mm}^{-1} \bm{f}_m, \tilde{Q}_{**}).
	\end{align}
\end{subequations}
Thereafter, we marginalize out all the latent variables to approximate the log marginal likelihood $\log p(\bm{y})$ as
\begin{equation} \label{eq_log_q(y)}
\begin{aligned}
\log q(\bm{y}) = -\frac{n}{2} \log(2 \pi) - \frac{1}{2} \log |\tilde{\bm{Q}}_{nn} + \bm{Q}_{nn} + \sigma^2_{\epsilon} \bm{I}| - \frac{1}{2} \bm{y}^{\mathsf{T}} (\tilde{\bm{Q}}_{nn} + \bm{Q}_{nn} + \sigma^2_{\epsilon} \bm{I})^{-1} \bm{y}.
\end{aligned}
\end{equation}
Through particular selections of $\tilde{\bm{Q}}_{nn}$, we can efficiently calculate the determinant $|\tilde{\bm{Q}}_{nn} + \bm{Q}_{nn} + \sigma^2_{\epsilon} \bm{I}|$ and the inversion $(\tilde{\bm{Q}}_{nn} + \bm{Q}_{nn} + \sigma^2_{\epsilon} \bm{I})^{-1}$ in~\eqref{eq_log_q(y)} by only depending on $\bm{K}^{-1}_{mm}$.

For instance, the subset-of-regressors (SoR)~\cite{smola2001sparse, silverman1985some, wahba1999bias} imposes $\tilde{\bm{Q}}_{nn} = \bm{0}$ and $\tilde{Q}_{**} = 0$ to the training and test conditionals such that
\begin{subequations} \label{eq_SoR_conditional}
	\begin{align}
	q_{\mathrm{SoR}}(\bm{f} | \bm{f}_m) &= \mathcal{N} (\bm{f}|\bm{K}_{nm} \bm{K}_{mm}^{-1} \bm{f}_m, \bm{0}),\\ q_{\mathrm{SoR}}(f_* | \bm{f}_m) &= \mathcal{N} (f_*|\bm{k}_{*m} \bm{K}_{mm}^{-1} \bm{f}_m, 0).
	\end{align}
\end{subequations}
This is equivalent to applying the Nystr\"{o}m approximation to both training and test data. With this consistent assumption, the SoR is equivalent to a degenerate\footnote{``degenerate'' means the GP employs a kernel with a finite rank.} GP with the rank (at most) $m$ kernel function $k_{\mathrm{SoR}}(\bm{x}_i, \bm{x}_j) = k(\bm{x}_i, \bm{X}_m) \bm{K}_{mm}^{-1} k^{\mathsf{T}}(\bm{x}_j, \bm{X}_m)$. However, due to the limited $m$ freedoms in $k_{\mathrm{SoR}}$, the SoR suffers from over-confident prediction variance when leaving the training data. In contrast to SoR, the deterministic training conditional (DTC)~\cite{ seeger2003fast,csato2002sparse} imposes $\tilde{\bm{Q}}_{nn} = \bm{0}$ but retains the exact test conditional as
\begin{subequations} \label{eq_DTC_conditional}
	\begin{align}
	q_{\mathrm{DTC}}(\bm{f} | \bm{f}_m) &= \mathcal{N} (\bm{f}|\bm{K}_{nm} \bm{K}_{mm}^{-1} \bm{f}_m, \bm{0}),\\ q_{\mathrm{DTC}}(f_* | \bm{f}_m) &= p(f_* | \bm{f}_m).
	\end{align}
\end{subequations}
Hence, the prediction mean $\mu_{\mathrm{DTC}}(\bm{x}_*)$ is the same as that of SoR, but the prediction variance $\sigma^2_{\mathrm{DTC}}(\bm{x}_*)$ is always larger than that of SoR. Notably, due to the inconsistent approximations in~\eqref{eq_DTC_conditional}, the DTC is not an exact GP. 

Moreover, the fully independent training conditional (FITC)~\cite{snelson2006sparse} imposes another independence assumption to remove the dependency among the latent variables $\{f_i\}_{i=1}^n$ such that the training conditional factorizes as
\begin{equation} \label{eq_q_FITC}
\begin{aligned}
q_{\mathrm{FITC}}(\bm{f} | \bm{f}_m) = \prod_{i=1}^n p(f_i | \bm{f}_m) = \mathcal{N} (\bm{f}|\bm{K}_{nm} \bm{K}_{mm}^{-1} \bm{f}_m, \mathrm{diag}[\bm{K}_{nn} - \bm{Q}_{nn}]),
\end{aligned}
\end{equation}
whereas the test conditional retains exact. It is found that the variances of $q_{\mathrm{FITC}}(\bm{f} | \bm{f}_m)$ are identical to that of $p(\bm{f} | \bm{f}_m)$ due to the diagonal correction term $\tilde{\bm{Q}}_{nn} = \mathrm{diag}[\bm{K}_{nn} - \bm{Q}_{nn}]$. Hence, compared to SoR and DTC which completely omit the uncertainty in~\eqref{eq_SoR_conditional} and~\eqref{eq_DTC_conditional}, FITC partially retains it, resulting in a closer approximation to the prior $p(\bm{f}, f_*)$. 

The extensions of FITC include for example the fully independent conditional (FIC)~\cite{quinonero2005unifying} which additionally applies the independence assumption to the test conditional $q(\bm{f}_* | \bm{f}_m)$ in order to obtain a degenerate GP with the covariance function $k_{\mathrm{FIC}}(\bm{x}_i, \bm{x}_j) = k_{\mathrm{SoR}}(\bm{x}_i, \bm{x}_j) + \delta_{ij} [k(\bm{x}_i, \bm{x}_j) - k_{\mathrm{SoR}}(\bm{x}_i, \bm{x}_j)]$, where $\delta_{ij}$ is the Kronecker's delta. Besides, the partially independent training conditional (PITC)~\cite{quinonero2005unifying} factorizes $q(\bm{f}|\bm{f}_m)$ over $M$ independent subsets (blocks) $\{ \mathcal{D}_i \}_{i=1}^M$, thus taking into account the joint distribution of $\bm{f}_i$ in each subset. However, it is argued in~\cite{snelson2007local} that though providing a closer approximation to $p(\bm{f} | \bm{f}_m)$, the PITC brings little improvements over FITC. 

Differently, the partially independent conditional (PIC)~\cite{snelson2007local} retains the conditional independence assumption, i.e., $\bm{f} \perp f_* | \bm{f}_m$, for all the blocks except the one containing the test point $\bm{x}_*$. Suppose that $\bm{x}_* \in \mathcal{D}_j$, the training conditional is
\begin{equation}
q_{\mathrm{PIC}}(\bm{f}, f_* | \bm{f}_m) = p(\bm{f}_j, f_*|\bm{f}_m) \prod_{i \neq j}^M p(\bm{f}_i | \bm{f}_m),
\end{equation}
which corresponds to an exact GP with $k_{\mathrm{PIC}}(\bm{x}_i, \bm{x}_j) = k_{\mathrm{SoR}}(\bm{x}_i, \bm{x}_j) + \psi_{ij} [k(\bm{x}_i, \bm{x}_j) - k_{\mathrm{SoR}}(\bm{x}_i, \bm{x}_j)]$, where $\psi_{ij} = 1$ when $\bm{x}_i$ and $\bm{x}_j$ belong to the same block; otherwise $\psi_{ij} = 0$. The PIC is regarded as a hybrid approximation since when we take all the block sizes to one, it recovers FIC; when we take the number of inducing points to zero, it recovers the pure local GP.

\subsection{Posterior approximations}
In contrast to prior approximations, posterior approximations, e.g., variational free energy (VFE)~\cite{titsias2009variational}, directly approximate the posterior $p(\bm{f}, \bm{f}_m | \bm{y})$ by a free variational distribution $q(\bm{f}, \bm{f}_m | \bm{y})$. The discrepancy between $q(\bm{f}, \bm{f}_m | \bm{y})$ and $p(\bm{f}, \bm{f}_m | \bm{y})$ can be quantified by the Kullback-Leibler (KL) divergence
\begin{equation}
\begin{aligned}
\mathrm{KL}(q(\bm{f}, \bm{f}_m | \bm{y}) || p(\bm{f}, \bm{f}_m | \bm{y})) =& \int q(\bm{f}, \bm{f}_m | \bm{y}) \log \frac{q(\bm{f}, \bm{f}_m | \bm{y})}{p(\bm{f}, \bm{f}_m, \bm{y})} d\bm{f} d\bm{f}_m + \log p(\bm{y}) \\
=& -F_q + \log p(\bm{y}).
\end{aligned}
\end{equation}
It is observed that minimizing the non-negative $\mathrm{KL}(q || p)$ is equivalent to maximizing $F_q$, since $\log p(\bm{y})$ is fixed w.r.t. $q(\bm{f}, \bm{f}_m | \bm{y})$. Thus, $F_q$ is called the lower bound of $\log p(\bm{y})$ or the variational free energy, which permits joint optimization of the variational parameters and hyperparameters.

Since we have $p(\bm{f}, \bm{f}_m | \bm{y}) = p(\bm{f} | \bm{f}_m) p(\bm{f}_m | \bm{y})$, the variational distribution factorizes similarly as $q(\bm{f}, \bm{f}_m | \bm{y}) = p(\bm{f} | \bm{f}_m) q(\bm{f}_m | \bm{y})$,
where $q(\bm{f}_m | \bm{y}) = \mathcal{N}(\bm{f}_m|\bm{m}, \bm{S})$. Fortunately, the optimal variational distribution $q^*(\bm{f}_m | \bm{y})$ can be found using the calculus of variations. Taking the derivative of $F_q$ w.r.t. $q(\bm{f}_m | \bm{y})$ to zero, we have
\begin{equation}
q^*(\bm{f}_m | \bm{y}) = \mathcal{N}(\bm{f}_m | \sigma^{-2}_{\epsilon} \bm{K}_{mm} \bm{\Sigma} \bm{K}_{mn} \bm{y}, \bm{K}_{mm} \bm{\Sigma} \bm{K}_{mm}),
\end{equation}
where $\bm{\Sigma} = (\bm{K}_{mm} + \sigma^{-2}_{\epsilon} \bm{K}_{mn} \bm{K}_{nm})^{-1}$. Inserting $q^*(\bm{f}_m | \bm{y})$ back into $F_q$, we have a ``collapsed'' bound which is independent of $q(\bm{f}_m | \bm{y})$ as
\begin{equation} \label{eq_F_vfe}
\begin{aligned}
F_{\mathrm{VFE}} =& -\frac{n}{2} \log(2 \pi) - \frac{1}{2} \log |\bm{Q}_{nn} + \sigma^2_{\epsilon} \bm{I}| \\
&- \frac{1}{2} \bm{y}^{\mathsf{T}} (\bm{Q}_{nn} + \sigma^2_{\epsilon} \bm{I})^{-1} \bm{y} - \frac{1}{2 \sigma^2_{\epsilon}} \mathrm{Tr} (\bm{K}_{nn} - \bm{Q}_{nn}) \\
=& \log q_{\mathrm{DTC}}(\bm{y}) - 0.5 \sigma^{-2}_{\epsilon} \mathrm{Tr} (\bm{K}_{nn} - \bm{Q}_{nn}).
\end{aligned}
\end{equation} 
It is found that compared to the log marginal likelihood of DTC, the VFE bound~\eqref{eq_F_vfe} involves an additional trace term $\mathrm{Tr} (\bm{K}_{nn} - \bm{Q}_{nn}) \ge 0$ which represents the total variance of the training conditional $p(\bm{f}|\bm{f}_m)$. Particularly, $\mathrm{Tr} (\bm{K}_{nn} - \bm{Q}_{nn}) = 0$ means that $\bm{f}_m = \bm{f}$ and we reproduce the full GP. Hence, the trace term helps VFE (i) guard against over-fitting; (ii) choose a good inducing set; and (iii) improve $F_{\mathrm{VFE}}$ with increasing $m$~\cite{titsias2009variational, titsias2009model, bauer2016understanding, matthews2016sparse}.

To further improve the scalability of VFE, unlike the bound~\eqref{eq_F_vfe} that ``integrates'' out the variational distribution, $q(\bm{f}_m | \bm{y})$ now is retained in the bound $F_q$ in order to obtain a full factorization over data points as~\cite{hensman2013gaussian}
\begin{equation} \label{eq_F_Q}
F_q = \left\langle \log p(\bm{y}|\bm{f}) \right\rangle_{p(\bm{f}|\bm{f}_m) q(\bm{f}_m | \bm{y})} - \mathrm{KL}(q(\bm{f}_m|\bm{y}) || p(\bm{f}_m)),
\end{equation}
where $\left\langle . \right\rangle_{q(.)}$ represents the expectation over the distribution $q(.)$. The newly organized bound $F_q$, which is looser than $F_{\mathrm{VFE}}$, has a key property that it can be written as the sum of $n$ terms because of the likelihood $p(\bm{y} | \bm{f}) = \prod_{i=1}^n p(y_i | f_i)$. Due to (i) the full factorization and (ii) the difficulty of optimizing the variational parameters $\bm{m}$ and $\bm{S}$ since they are defined in a non-Euclidean space, one can employ the Stochastic Variational Inference (SVI)~\cite{hoffman2013stochastic} to infer the variational parameters in the natural gradient space via efficient stochastic gradient descent (SGD) algorithms~\cite{zeiler2012adadelta, kingma2014adam}, resulting in the greatly reduced complexity of $\mathcal{O}(bm^3)$ where $b$ is the mini-batch size in SGD. Therefore, the stochastic variational GP (SVGP) adopts the SGD to train a sparse GP at \textit{any time} with a \textit{small batch} of the training data in each iteration~\cite{hoang2015unifying}.\footnote{Unlike the traditional conjugate gradient descent (CGD) algorithm which takes all the training data to calculate the gradients, the SGD only takes $b$ ($b \ll n$) out of the $n$ data points to calculate the gradients in each iteration, thus enhancing large-scale learning.}

\section{Local approximations}
\label{sec_local}
Following the idea of D\&C, the simplest local approximation is the pure local GPs which first partition the training data into $M$ subsets for example through clustering techniques, and then train a GP expert on each subset using individual parameters. The D\&C idea scales local approximations up to arbitrary dataset due to the straightforward parallel/distributed fashion. Besides, in comparison to the global approximations, the naive local approximation is capable of describing non-stationary features (local patterns) for improving predictions at the cost of however suffering from discontinuous predictions, inaccurate uncertainties and local over-fitting. 

To smooth and boost predictions while retaining computation gains from D\&C, the local aggregations, which are inspired by the idea of model averaging, were presented to aggregate the predictions from multiple experts. Particularly, given a partition $\{\mathcal{D}_i \}_{i=1}^M$ of $\mathcal{D}$, the aggregation methods follow the product rule~\cite{hinton2002training}, i.e., assuming independent GP experts, such that the marginal likelihood $p(\bm{y})$ factorizes as
\begin{equation}
p(\bm{y}|\bm{X},\bm{\theta}) \approx \prod_{i=1}^M p_i(\bm{y}_i|\bm{X}_i,\bm{\theta}_i),
\end{equation}
where $\bm{\theta}_i$ is a vector comprising the hyperparameters for the \textit{i}th expert $\mathcal{M}_i$, and $\bm{\theta} = \{\bm{\theta}_1, \cdots, \bm{\theta}_M \}$. Similar to the sparse approximations, the aggregations reduce the time complexity of full GP to $\mathcal{O}(nm_0^2)$ when all the experts have an equal training size $m_0 = n/M$, but requiring no additional inducing points or variational parameters.

After training the experts $\{ \mathcal{M}_i \}_{i=1}^M$ on the relevant subsets $\{\mathcal{D}_i\}_{i=1}^M$, we combine their predictions together at the test point $\bm{x}_*$ by employing some aggregation criteria. For instance, under the independence assumption, the product-of-experts (PoE)~\cite{hinton2002training,cao2014generalized, chen2009bagging, okadome2013fast, van2015optimally, van2017cluster} conducts the aggregation as
\begin{equation} \label{eq_PoE}
p_{\mathcal{A}}(y_*|\mathcal{D}, \bm{x}_*) = \prod_{i=1}^M p_i^{\beta_i}(y_*|\mathcal{D}_i, \bm{x}_*),
\end{equation}
where $\beta_i$ is a weight quantifying the contribution of $\mathcal{M}_i$ at $\bm{x}_*$. The product of Gaussian distributions in~\eqref{eq_PoE} results in another Gaussian distribution with the mean and variance analytically expressed as
\begin{subequations}
	\begin{align}
	\mu_{\mathcal{A}}(\bm{x}_*) = & \sigma^2_{\mathcal{A}}(\bm{x}_*) \sum_{i=1}^M \beta_i \sigma^{-2}_i(\bm{x}_*) \mu_i(\bm{x}_*), \label{eq_mu_PoE} \\
	\sigma^{-2}_{\mathcal{A}}(\bm{x}_*) =& \sum_{i=1}^M \beta_i \sigma^{-2}_i(\bm{x}_*), \label{eq_s2_PoE}
	\end{align}
\end{subequations}
where $\mu_i(\bm{x}_*)$ and $\sigma^2_i(\bm{x}_*)$ are respectively the prediction mean and variance of $\mathcal{M}_i$ at $\bm{x}_*$. By taking a constant weight $\beta_i = 1$, the original PoE quantifies the contribution of each expert by the prediction precision~\cite{hinton2002training}. However, the naive sum of experts' prediction precisions in~\eqref{eq_s2_PoE} will make the aggregated variance vanish quickly with increasing $M$, i.e., producing seriously over-confident prediction variance~\cite{ liu2018generalized,szabo2017asymptotic}. Hence, the generalized PoE (GPoE)~\cite{cao2014generalized} introduces a varying weight $\beta_i = 0.5(\log \sigma^2_{**} - \log \sigma_i^2(\bm{x}_*))$, which is defined as the difference in the differential entropy between the prior and the posterior, to weaken the votes of those poor experts with large uncertainty. This flexible weight, however, produces an explosive prediction variance when $\bm{x}_*$ is far away from $\bm{X}$~\cite{liu2018generalized}. To address this issue, we could either impose a constraint $\sum_{i=1}^M \beta_i = 1$~\cite{cao2014generalized} or simply employ $\beta_i = 1/M$~\cite{deisenroth2015distributed}.

To improve the performance of PoEs, the Bayesian committee machine (BCM)~\cite{deisenroth2015distributed,tresp2000bayesian, liu2018generalized,mair2018distributed} additionally takes the prior $p(y_*|\bm{x}_*)$ into account and imposes a conditional independence assumption $p(\bm{y}|f_*,\bm{X}) = \prod_{i=1}^M p(\bm{y}_i | f_*,\bm{X}_i)$. Consequently, the aggregated predictive distribution derived from the Bayes rule is
\begin{equation} \label{eq_BCM}
p_{\mathcal{A}}(y_*|\mathcal{D}, \bm{x}_*) = \frac{\prod_{i=1}^M p_i^{\beta_i}(y_*|\mathcal{D}_i, \bm{x}_*)}{p^{\sum_{i=1}^M \beta_i - 1}(y_*|\bm{x}_*)},
\end{equation}
with the prediction mean and variance analytically given as
\begin{subequations}
	\begin{align}
	\mu_{\mathcal{A}}(\bm{x}_*) = & \sigma^2_{\mathcal{A}}(\bm{x}_*) \sum_{i=1}^M \beta_i \sigma^{-2}_i(\bm{x}_*) \mu_i(\bm{x}_*), \\
	\sigma^{-2}_{\mathcal{A}}(\bm{x}_*) =& \sum_{i=1}^M \beta_i \sigma^{-2}_i(\bm{x}_*) + \left( 1 - \sum_{i=1}^M \beta_i \right)\sigma^{-2}_{**}. \label{eq_s2_BCM}
	\end{align}
\end{subequations}
Compared to~\eqref{eq_s2_PoE}, the prior correlation in~\eqref{eq_s2_BCM} helps BCMs recover the GP prior when leaving the training data. The original BCM~\cite{tresp2000bayesian} takes $\beta_i = 1$, and the newly developed robust BCM (RBCM)~\cite{deisenroth2015distributed} employs the varying $\beta_i$ like GPoE in order to produce robust predictions within $\bm{X}$. The BCMs however are found to suffer from weak experts when leaving $\bm{X}$~\cite{deisenroth2015distributed, liu2018generalized}.

Regarding the model capability, the PoEs in~\eqref{eq_PoE} allow the GP experts to own individual parameters in order to capture non-stationary features, whereas the BCMs in~\eqref{eq_BCM} cannot due to the shared prior $p(y_*|\bm{x}_*)$ across experts. Besides, for the BCMs using experts with shared hyperparameters, the direct aggregation of $\{p_i(y_*|\mathcal{D}_i, \bm{x}_*)\}_{i=1}^M$ induces seriously over-confident prediction variance with increasing $n$~\cite{liu2018generalized}. To alleviate this issue, we could use the newly proposed generalized RBCM framework~\cite{liu2018generalized}; or simply, we aggregate $\{p_i(f_*|\mathcal{D}_i, \bm{x}_*)\}_{i=1}^M$ instead of $\{p_i(y_*|\mathcal{D}_i, \bm{x}_*)\}_{i=1}^M$, and finally add the estimated noise variance~\cite{deisenroth2015distributed}.

\begin{table}
	\caption{Comparison of the space and time complexity of representative global and local scalable GPs, where $m$ is the inducing size for sparse approximations and $m_0$ is the training size of each expert for local aggregations.}
	\label{Tab_complexity}
	\centering
	\begin{tabular}{lccccc}
		\hline
		Model & Storage & Training & Test \\
		\hline
		SoR, DTC, FITC, VFE & $\mathcal{O}(nm)$  & $\mathcal{O}(nm^2)$ & $\mathcal{O}(m^2)$ \\
		SVGP & $\mathcal{O}(bm^2)$  & $\mathcal{O}(bm^3)$ & $\mathcal{O}(m^2)$  \\
		\hline
		(G)PoE, (R)BCM & $\mathcal{O}(nm_0)$ & $\mathcal{O}(nm_0^2)$ & $\mathcal{O}({nm_0})$ \\
		\hline
	\end{tabular}
\end{table}

Finally, Table~\ref{Tab_complexity} summarizes the training and test time complexity of representative global and local scalable GPs. It is found that if $m = m_0$, all the scalable GPs except SVGP own the same training complexity. Besides, the local aggregations have a higher test complexity since they need the predictions of all the experts at $\bm{x}_*$. Finally, note that the computations in the scalable GPs listed above can be sped up through distributed/parallel computing, see~\cite{deisenroth2015distributed,gal2014distributed, dai2014gaussian}.

\section{Numerical experiments}
\label{sec_exps}
This section first employs several toy examples to illustrate the methodological characteristics of global and local scalable GPs, and then applies some of them to five real-world datasets with up to 250K data points. The goal of this comparative study is to enhance the understanding of representative scalable GPs and investigate their usability by addressing the issues as: (i) what are the features of these scalable GPs and in what scenarios they are useful? (ii) are the scalable GPs controllable to model parameters? and (iii) are the scalable GPs robust to various initializations of hyperparameters?

In the comparative study below, we implement the scalable GPs based on the GPML toolbox\footnote{\url{http://www.gaussianprocess.org/gpml/code/matlab/doc/}}, the GPstuff toolbox\footnote{\url{https://github.com/gpstuff-dev/gpstuff}} and the GPy toolbox\footnote{\url{https://github.com/SheffieldML/GPy}}.
Before model training, the data pre-processing is performed by normalizing $\bm{y}$ and each column of $\bm{X}$ to $\mathcal{N}(0,1)$. In modeling, we employ the SE kernel in~\eqref{eq_SE_Kernel}, and initialize the length-scales $l_1, \cdots, l_d$ as 0.5, the signal variance $\sigma^2_f$ as 1.0, and the noise variance $\sigma^2_{\epsilon}$ as 0.1. All the scalable GPs except SVGP employ the CGD for inference with the maximum number of iterations as 100. The particular SVGP employs the Adadelta SGD algorithm~\cite{zeiler2012adadelta} for inference with the step rate as 0.1, the momentum as 0.9, and the maximum number of iterations as 1000. All the codes are executed on a personal computer with four 3.70 GHz cores and 16 GB RAM. 

Finally, given $n_*$ test points $\{ \bm{X}_*, \bm{y}_* \}$, we assess the prediction accuracy of scalable GPs using the standardized mean square error (SMSE) defined as
\begin{equation}
\mathrm{SMSE} = \frac{\sum_{j=1}^{n_*} \left(y_{*j} - \mu_{*j} \right)^2}{n_* \times \mathrm{var}(\bm{y})}.
\end{equation}
The SMSE criterion quantifies the discrepancy between the predictions and the exact function values; particularly, it equals to one when the model always predicts the mean of $\bm{y}$. Besides, to quantify the quality of predictive distribution that considers both prediction mean and variance, we employ the mean standardized log loss (MSLL) defined as
\begin{equation}
\mathrm{MSLL} = \frac{1}{n_*} \sum_{j=1}^{n_*} \left[\log \mathcal{N}(y_{*j} |\overline{\bm{y}}, \mathrm{var}(\bm{y}))-\log p(y_{*j} | \mathcal{D}, \bm{x}_{*j}) \right],
\end{equation}
where $\log p(y_{*j} | \mathcal{D}, \bm{x}_{*j}) = -0.5\left[ \log(2 \pi \sigma^2_{*j}) + (y_{*j} - \mu_{*j})^2/\sigma^2_{*j} \right]$. The MSLL will be negative for high quality models, and particularly, it will be zero when the model always predicts the mean and variance of $\bm{y}$.

\subsection{Toy examples} \label{sec_toy}
\subsubsection{Characteristics of sparse approximations}
\begin{figure}[hbt!] 
	\centering
	\includegraphics[width=0.9\textwidth]{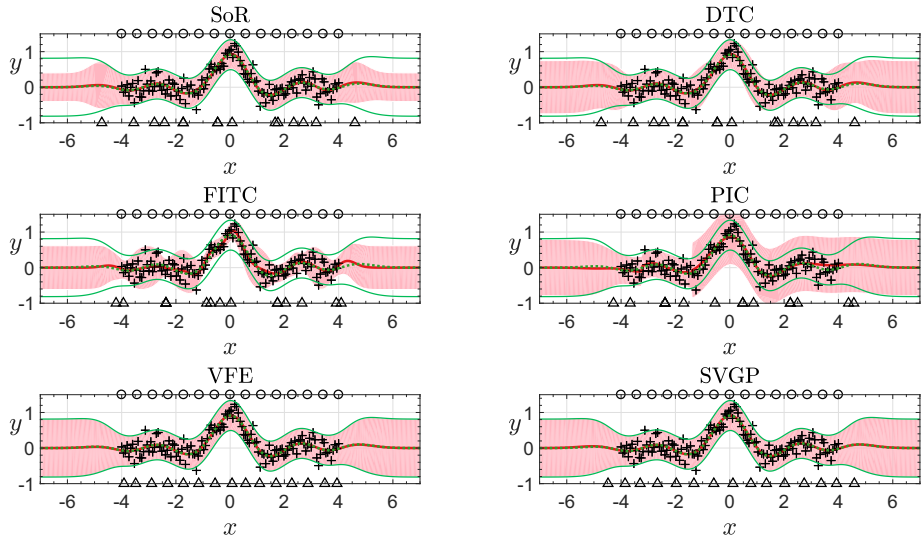}
	\caption{A toy example to illustrate the characteristics of various sparse approximations. The crosses represent the training points. The green dot curve represents the prediction mean of full GP. The two green curves represent 95\% confidence interval of the full GP prediction. The red curve represents the prediction mean of a sparse approximation. The shadow region represents 95\% confidence interval of the sparse prediction. The top circles and bottom triangles represent the positions of initial and optimized inducing points, respectively.}
	\label{Fig_sparse_toy} 
\end{figure}

This section attempts to study various sparse approximations via a toy example expressed as
\begin{equation} \label{eq_toy1}
y(x) = \mathrm{sinc}(x) + \epsilon, \quad x \in [-4,4],
\end{equation}
where $\epsilon = \mathcal{N}(0, 0.04)$. We randomly draw 120 training points from this generative function; besides, we generate 300 test points in $[-7,7]$. The sparse approximations include SoR, DTC, FITC, PIC, VFE and SVGP. As for model configurations, we choose 15 initial inducing points equally spaced in $[-4, 4]$; particularly, we partition the training data into $M = 10$ disjoint subsets for the PIC approximation; we use the batch size of $b=30$ for SVGP. Fig.~\ref{Fig_sparse_toy} depicts the predictions of the six sparse approximations, respectively, on the toy example. For the purpose of comparison, the results of full GP are involved in the figure.

\begin{figure}[hbt!] 
	\centering
	\includegraphics[width=1.0\textwidth]{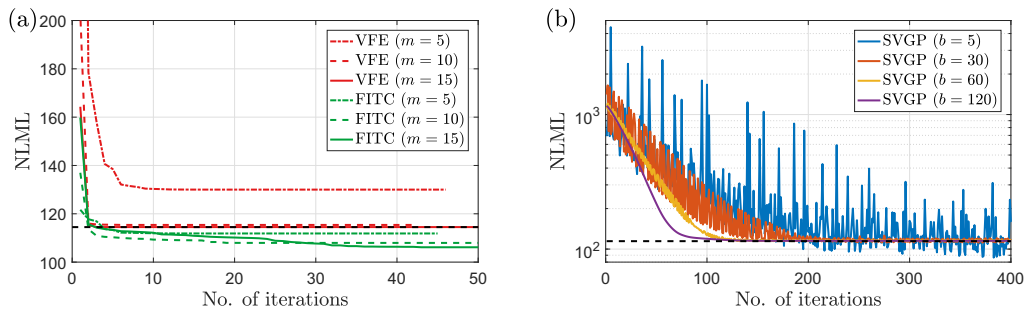}
	\caption{The convergence curves of (a) VFE and FITC using CGD with different inducing sizes, and (b) SVGP using SGD with different batch sizes on the toy example.}
	\label{Fig_VFE_FITC_SVGP_convergence_toy} 
\end{figure}

It is first observed that the two posterior approximations, VFE and its variant SVGP, provide the best approximation to the full GP, since they directly approximate the posterior with no modification to the joint prior. If we gradually increase the inducing size $m$, they will finally converge to the full GP. To verify this, Fig.~\ref{Fig_VFE_FITC_SVGP_convergence_toy}(a) depicts the convergence curves of VFE with different $m$ values. The horizontal axis represents the number of optimization iterations, and the vertical axis represents the negative log marginal likelihood (NLML). The black dash line indicates the converged NLML value of full GP. It is observed that when using a small inducing set ($m=5$), the VFE converges with a larger NLML value than that of full GP; but with the increase of $m$, the NLML of VFE quickly converges to that of full GP. Note that because of the simplicity of this toy example, the VFE with $m=15$ has provided a very close approximation to the full GP.

The differences between VFE and SVGP are that (i) SVGP employs a less tight bound~\eqref{eq_F_Q} defined in an augmented probabilistic space;\footnote{Due to the explicit variational distribution $q(\bm{f}_m|\bm{y})$ in~\eqref{eq_F_Q}, the SVGP should consider $m+m(m+1)/2$ additional variational parameters.} and (ii) SVGP employs the SGD for optimization. Fig.~\ref{Fig_VFE_FITC_SVGP_convergence_toy}(b) shows the convergence curves of SVGP using different batch sizes for SGD. It is observed that compared to the deterministic CGD, the SGD (i) produces many fluctuations with a small batch size; and (ii) converges more slowly even by using $b = 120$, because of the relaxed bound $F_q$~\eqref{eq_F_Q} and the huge parameter space. But the superiority of SGD for big data is that (i) it greatly reduces the computational complexity by running in a batch mode; and (ii) it can achieve better solutions for complicated functions by easily escaping from local optima and saddle points. For example, the SVGP with $b=5$ in Fig.~\ref{Fig_VFE_FITC_SVGP_convergence_toy}(b) finds some smaller NLML values than that of full GP.

As for the four prior approximations, it is observed in Fig.~\ref{Fig_sparse_toy} that the SoR produces severely over-confident prediction variance when leaving $\bm{X}$. This is because the SoR imposes too restrictive assumptions to the training and test data in~\eqref{eq_SoR_conditional}. Though equipped with the same prediction expressions to that of VFE, the DTC produces poorer predictions due to the restrictive marginal likelihood. Different from SoR and DTC, the FITC is capable of capturing the heteroscedastic noise. Let us look inside the prediction variance of FITC
\begin{equation*}
\sigma^2(\bm{x}_*) = k_{**} - Q_{**} + \bm{k}_{*m} \bm{\Sigma}_{mm} \bm{k}_{m*},
\end{equation*}
where $\bm{\Sigma}_{mm} = (\bm{K}_{mm} + \bm{K}_{mn} \bm{\Lambda}^{-1} \bm{K}_{nm})^{-1}$ and $\bm{\Lambda} = \mathrm{diag}[\bm{K}_{nn} - \bm{Q}_{nn}] + \sigma^2_{\epsilon} \bm{I}_n$. It is found that the diagonal term $\mathrm{diag}[\bm{K}_{nn} - \bm{Q}_{nn}]$, which represents the input-dependent variances of the training conditional $p(\bm{f}|\bm{f}_m)$, enables FITC to capture the heteroscedasticity of noise at the cost of (i) producing an invalid estimation (nearly zero) of the noise variance $\sigma^2_{\epsilon}$, (ii) worsening the accuracy of prediction mean, and (iii) producing overlapped inducing points~\cite{bauer2016understanding}. Finally, the hybrid PIC approximation, which is capable of capturing local patterns, is found to produce discontinuous predictions and conservative variances.

As discussed before, the posterior approximations VFE and SVGP can be regarded as the approximation to full GP. But this is not the case for the four prior approximations. Prior approximations seek to achieve good prediction accuracy at a low computational cost, rather than faithfully converging to the full GP with increasing $m$. This can be reflected by two observations: (i) some of the optimized inducing points in Fig.~\ref{Fig_sparse_toy} overlap with each other, especially for FITC;\footnote{The reason for FITC to produce overlapped inducing points has been theoretically analyzed in~\cite{bauer2016understanding}.} and (ii) rather than converging to the NLML of full GP with the increase of $m$, the FITC tends to produce a different, smaller NLML value in Fig.~\ref{Fig_VFE_FITC_SVGP_convergence_toy}(a).

\subsubsection{Characteristics of local approximations}
\begin{figure}[hbt!] 
	\centering
	\includegraphics[width=1.0\textwidth]{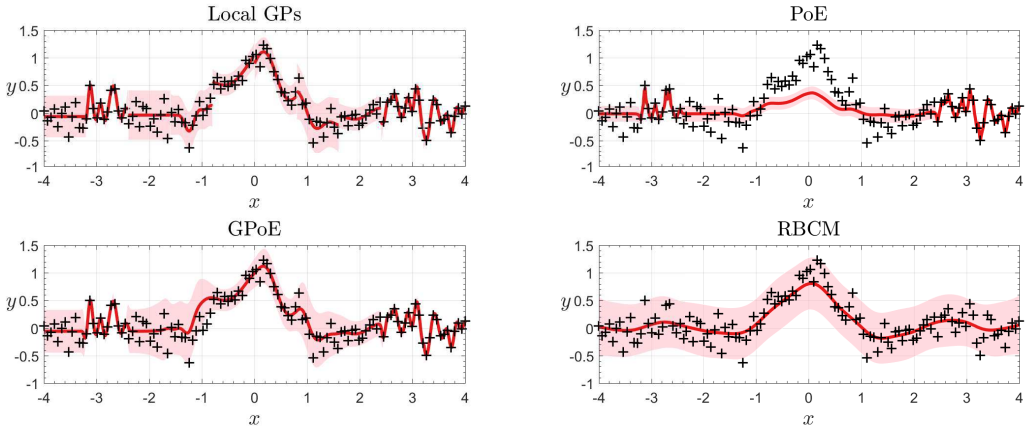}
	\caption{A toy example to illustrate the characteristics of various local approximations.}
	\label{Fig_local_toy} 
\end{figure}

Different from the sparse approximations which capture long-term spatial correlations based on the global inducing set, the local approximations focus on subspace learning via local experts, thus capturing local patterns (non-stationary features). As an illustration, we apply the pure local GPs to the toy example~\eqref{eq_toy1} in Fig.~\ref{Fig_local_toy}. Particularly, we partition the 120 training points into $M=10$ local subsets and train individual GP experts. It is found that the naive local approximation captures local patterns at the cost of (i) producing discontinuity on the boundaries of sub-regions, and (ii) risking over-fitting in some local regions.

The first issue could be addressed by the model averaging strategies like PoE and BCM. As shown in Fig.~\ref{Fig_local_toy}, the PoE and GPoE yield continuous predictions. But the original product rule makes PoE produce over-confident prediction variance and deteriorated prediction mean, which are not preferred in practice.  Instead, the GPoE employs a varying weight $\beta_i$ to weaken the votes of poor experts, resulting in sensible prediction mean and variance.

The second issue however is directly inherited by PoE and GPoE, since they have no mechanism to avoid local over-fitting. To alleviate the over-fitting issue, we could increase the training size for each expert (i.e., small $M$ value) in order to take into account the long-term spatial correlation at the cost of degrading the capability of capturing local patterns and increasing the complexity. Besides, the hybrid approximations~\cite{snelson2007local, vanhatalo2010approximate, lee2017hierarchically} may be a promising solution to this issue, since they inherit the advantages of both global and local approximations.\footnote{The current hybrid approximations, e.g., the PIC in Fig.~\ref{Fig_sparse_toy}, however still suffer from the discontinuity issue.} Finally, another alternative way to guard against local over-fitting is to sharing hyperparameters across experts, like the RBCM in Fig.~\ref{Fig_local_toy}. It is observed that the RBCM has conservative prediction variances. Note that though sharing the hyperparameters, the local structure itself could help RBCM capture local patterns, which will be shown in next section.

\subsubsection{Global vs. local approximations}
Here we take the time-series \textit{solar} dataset~\cite{hensman2016variational} which contains 391 data points with \textit{quick-varying} features for comparing global and local approximations. The study attempts to investigate whether global and local approximations could handle complicated tasks using limited computational resources.

Among the four local approximations, we take the RBCM for example and use the \textit{k}-means technique to partition the training data into $M=10$ disjoint subsets, resulting in approximately 39 data points for each expert. Among the six global approximations, we employ the VFE and use $m=40$ inducing points. Particularly, in order to model the quick-varying features in this dataset, we initialize the length-scales in the SE kernel~\eqref{eq_SE_Kernel} with a small value of 0.0442 after data normalization.\footnote{In the original input space $[1600,2000]$ of the \textit{solar} dataset, the initial length-scale is equal to 5, which is a very small value.}

\begin{figure}[hbt!] 
	\centering
	\includegraphics[width=0.7\textwidth]{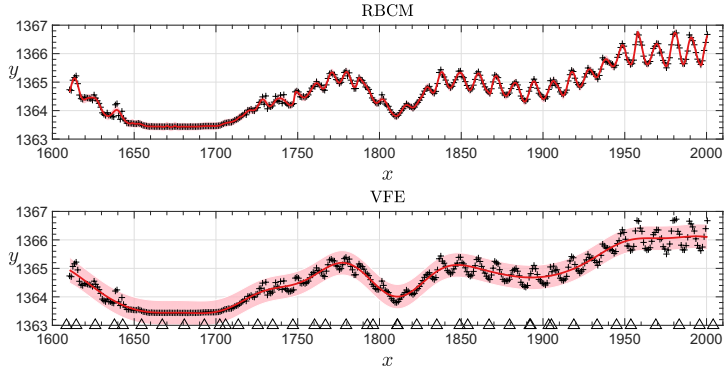}
	\caption{A \textit{solar} example involving quick-varying features to illustrate the characteristics of RBCM and VFE.}
	\label{Fig_RBCMvsVFE} 
\end{figure}

Fig.~\ref{Fig_RBCMvsVFE} depicts the modeling results of RBCM and VFE on the \textit{solar} dataset. It turns out that though sharing the hyperparameters across experts, the RBCM captures the quick-varying features successfully. This is because the \textit{localized structure} helps RBCM equipped with \textit{local attention} to take into account the local patterns when estimating the shared hyperparameters. On the contrary, even by using such small initial length-scales, the VFE fails to capture the quick-varying features due to the small set of \textit{global} inducing points. The performance of VFE indeed could be improved by increasing the number of inducing points, which however becomes unattractive in terms of computational complexity.

Besides, the local attention mechanism may help improve the robustness of RBCM to various settings. To verify this, we study an extreme case wherein the length-scales in the SE kernel are initialized as 2.0 which in the original space is 226. By simply increasing the number of experts to $M=40$, the RBCM again is enabled to capture the quick-varying feature at, interestingly, lower computational cost.  But it is notable that the increase of $M$ does not always correspond to good predictions, since practical datasets often do not follow the \textit{iid} noise assumption. That means too much localized experts for RBCM may degrade the generalization capability (indicated by severely over-confident prediction variance), which will be observed in next section.

\subsection{Real-world datasets}
\label{sec_real_world}
This section seeks to assess the representative global and local scalable GPs on five real-world datasets with different characteristics, e.g., regular/clustering inputs, noise/noiseless observations, and homoscedastic/heteroscedastic noise, see Table~\ref{Tab_realWorld}. The \textit{airfoil} dataset~\cite{Dua:2017} comprises different sizes of NACA 0012 airfoils at various wind tunnel speeds and angles of attack, and the output is the scaled sound pressure level. The \textit{protein} dataset~\cite{Dua:2017} describes the physicochemical properties of the protein tertiary structure. The \textit{sarcos} dataset~\cite{rasmussen2006gaussian} describes the inverse kinematics of a robot arm. The \textit{chem} dataset~\cite{malshe2007theoretical} concerns the physical simulations relating to electron energies in molecules. Finally, the \textit{sdss} dataset~\cite{almosallam2016gpz} comes from the Sloan Digital Sky Survey’s 12th Data Release. Note that each configuration of $n$ and $n_*$ in Table~\ref{Tab_realWorld} has ten random instances in order to comprehensively evaluate the performance of scalable GPs.

\begin{table}
	\caption{Characteristics of five real-world datasets.}
	\label{Tab_realWorld}
	\centering
	\begin{tabular}{lcccccc}
		\hline
		dataset & $d$ & $n$ & $n_*$ & remark \\\hline
		\textit{airfoil} & 5  & 1,200 & 303 & regular inputs \\
		\textit{protein} & 9  & 35,000 & 10,730 & clustering inputs  \\
		\textit{sarcos} & 21 & 40,000 & 8,933 & nearly noiseless \\
		\textit{chem} & 15 & 60,000 & 11,969 & heteroscedastic, tiny noise \\
		\textit{sdss} & 10  & 250,000 & 50,000 & heteroscedastic noise \\
		\hline
	\end{tabular}
\end{table}

Among the six global scalable GPs in Fig.~\ref{Fig_VFE_FITC_SVGP_convergence_toy}, we select the VFE and SVGP due to the high approximation quality, and the FITC due to the capability of capturing heteroscedastic noise; the SoR, DTC and PIC are not included due to their poor or discontinuous predictions. Among the four local scalable GPs in Fig.~\ref{Fig_local_toy}, we choose the RBCM trained on experts with shared hyperparameters and the GPoE trained on experts with individual hyperparameters; the pure local GPs and PoE are not considered due to their less competitive performance.

\subsubsection{Comparison of global and local scalable GPs}
\label{sec_comp_global_local}
We first study the scalability and capability of global and local scalable GPs. Table~\ref{Tab_model_parameters_realWorld} offers the model parameters of scalable GPs, including the inducing size $m$, the batch size $b$ and the number $M$ of experts, for the five real-world datasets. During the comparison study, the inducing points are initialized by the centroids of clusters partitioned by the \textit{k}-means technique; the disjoint local subsets are partitioned by the \textit{k}-means technique as well. We choose the model parameters in Table~\ref{Tab_model_parameters_realWorld} such that these scalable GPs have comparable running time. The data pre-processing and optimization configurations are consistent to that in Section~\ref{sec_toy}.

\begin{table}
	\caption{Model parameters of scalable GPs for the five real-world datasets. $m$ is the inducing size for VFE, SVGP and FITC, $b$ is the batch size for SVGP, and $M$ is the number of experts for GPoE and RBCM.}
	\label{Tab_model_parameters_realWorld}
	\centering
	\begin{tabular}{lccccc}
		\hline
		Parameter & \textit{airfoil} & \textit{protein} & \textit{sarcos} & \textit{chem} & \textit{sdss}  \\
		\hline 
		$m$ & 60 & 400 & 400 & 300 & 250 \\
		$b$ & 50 & 2,500 & 2,500 & 3,000 & 9,000 \\
		$M$ & 20 & 35 & 50 & 80 & 500 \\
		\hline
	\end{tabular}
\end{table}

\begin{figure}[hbt!] 
	\centering
	\includegraphics[width=1.0\textwidth]{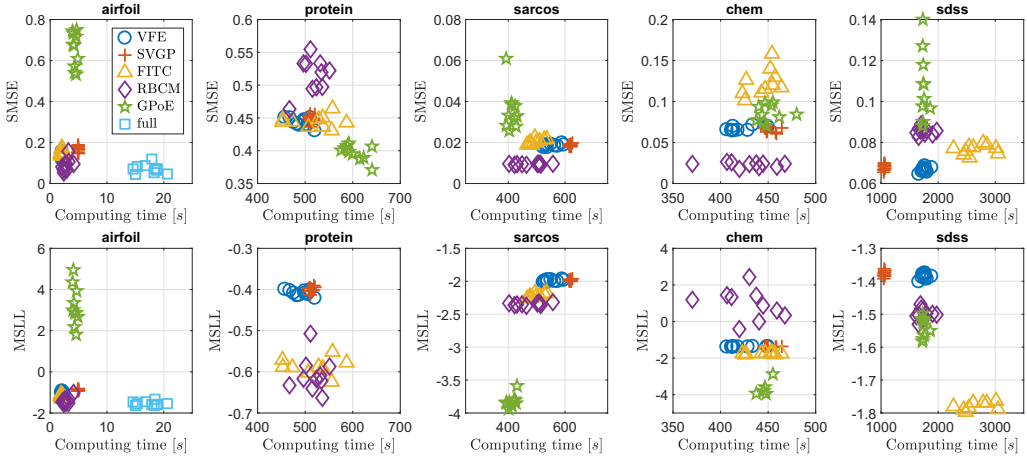}
	\caption{Comparative results of VFE, SVGP, FITC, RBCM and GPoE on the five real-world datasets. The results of full GP are provided for the \textit{airfoil} dataset with $n=1200$. Note that the GPoE has no MSLL symbols on the \textit{protein} dataset since it produces invalid MSLL values in all the ten runs. Similarly, we only show five success runs of GPoE in terms of MSLL on the \textit{chem} dataset.}
	\label{Fig_allDatasets_smse_msll} 
\end{figure}

Fig.~\ref{Fig_allDatasets_smse_msll} depicts the modeling results of five scalable GPs over ten runs on the five datasets in terms of SMSE and MSLL. The horizontal axis represents the sum of training and predicting time. Note that due to the small training size, we include the results of full GP on the \textit{airfoil} dataset for comparison. 

As for the three global scalable GPs, the VFE and SVGP produce similar results since they are derived in the same posterior approximation framework. However, the SVGP is more potential in terms of efficiency for large-scale learning since it allows using the SGD optimization, see the results on the \textit{sdss} dataset. Different from VFE and SVGP which follow a constant noise assumption, the FITC is capable of describing possible heteroscedastic noise variances, which are indicated by the smaller MSLL values in the figure. But this superiority of FITC comes at the cost of worsening the accuracy of prediction mean, resulting in larger SMSE values.

\begin{figure}[hbt!] 
	\centering
	\includegraphics[width=0.6\textwidth]{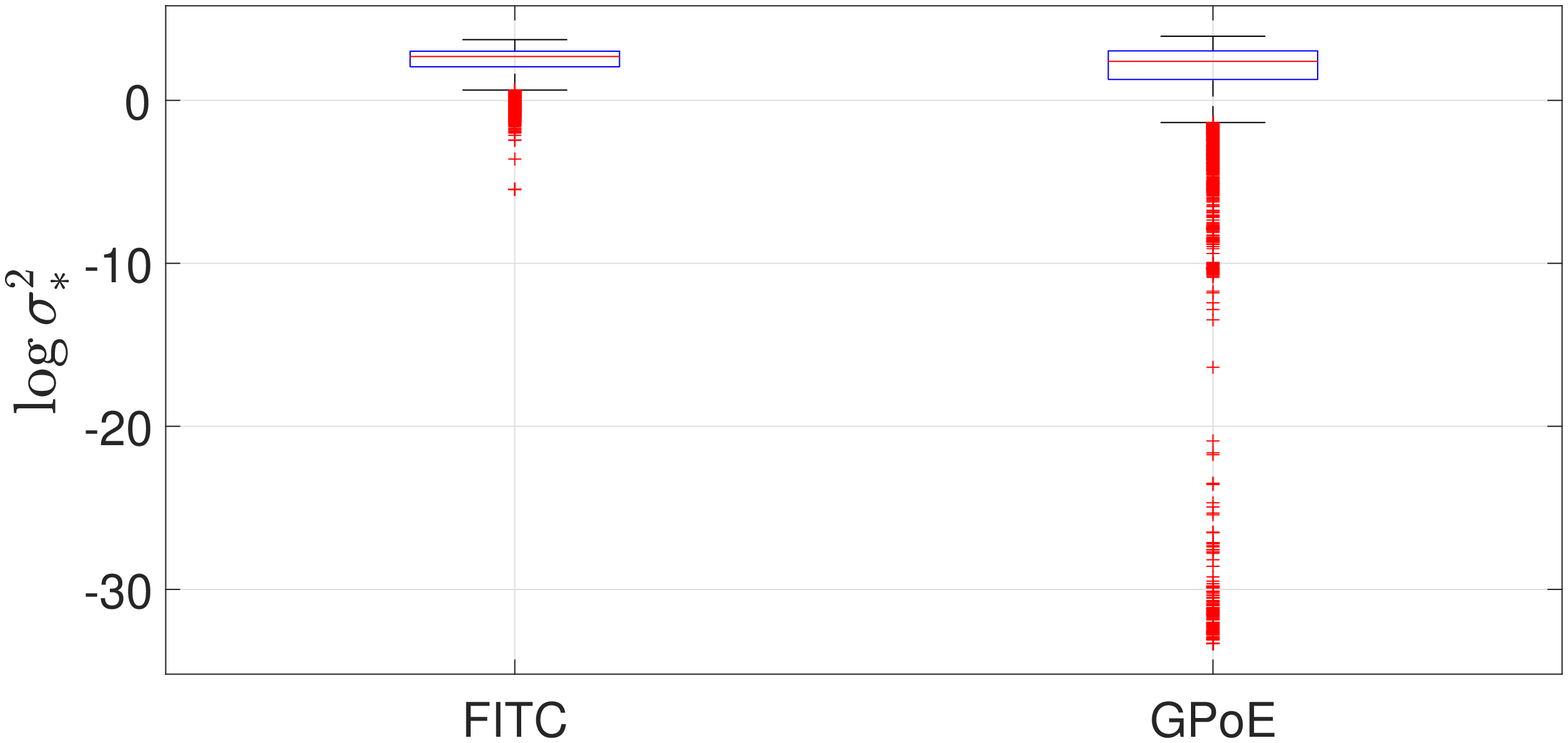}
	\caption{The boxplots of $\log \sigma^2_*$ estimated by FITC and GPoE in a run on the \textit{protein} dataset.}
	\label{Fig_s2_FITCvsGPoE_protein} 
\end{figure}

As for the two local scalable GPs, due to the individual experts, the GPoE captures better predictive distributions on three out of the five datasets, and produces more accurate predictions on the \textit{protein} dataset. However, it is observed that the GPoE yields invalid prediction variances on the \textit{protein} and \textit{chem} datasets due to the existence of local over-fitting. For instance, Fig.~\ref{Fig_s2_FITCvsGPoE_protein} depicts the $\log \sigma^2_*$ values estimated respectively by FITC and GPoE, since both of which can capture the heteroscedasticity in noise variance, on the \textit{protein} dataset. It is observed that in comparison to FITC, the GPoE produces many extremely small prediction variances due to local over-fitting. These extreme prediction variances of GPoE, as low as nearly $e^{-35}$, in turn bring invalid MSLL with the value up to $6.59 \times 10^{10}$. Besides, it is observed that compared to the RBCM, the individual treatment of GP experts in GPoE often induces ill-conditioned kernel matrix, especially in the scenario with many experts. This prohibits the model inference. The local over-fitting and the ill-conditioned phenomenon of GPoE could be alleviated by using a small $M$, i.e., large experts that can take into account the the spatial correlations. But this would degrade the model capability and improve the complexity of GPoE. Hence, in practice we prefer RBCM, unless there exist some tricks to guard against local over-fitting and ill-conditioned kernel matrix in GPoE.

Finally, the comparison between global and local scalable GPs shows that the local attention mechanism enables RBCM to produce smaller MSLL values than that of VFE and SVGP in four out of the five datasets, and smaller SMSE values in three out of the five datasets.

\subsubsection{Impact of model parameters}
It is found that the model parameters, e.g., the inducing size $m$ and the number $M$ of experts, affect the performance of scalable GPs. More inducing points bring better distillation of training data for sparse approximations, while more experts enhance the localization by taking into account more local patterns for local aggregations. Hence, this section seeks to run global and local scalable GPs using various model parameters in order to investigate their controllability. i.e., explicitly controlling the model performance by tuning $m$ or $M$.  

\begin{figure}[hbt!] 
	\centering
	\includegraphics[width=1.0\textwidth]{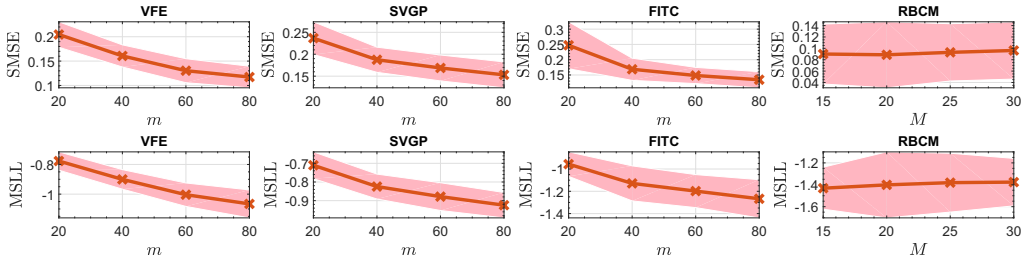}
	\caption{Impact of varying inducing size $m$ or number $M$ of experts on the performance of scalable GPs on the \textit{airfoil} dataset. The shaded region represents two times the standard deviation over ten runs.}
	\label{Fig_varyingParas_airfoil} 
\end{figure}

Fig.~\ref{Fig_varyingParas_airfoil} depicts the modeling results of scalable GPs using different $m$ or $M$ values on the \textit{airfoil} dataset. The results of scalable GPs on the remaining four datasets are provided in the Appendix. Note that the GPoE is not included in the comparison due to the unstable performance induced by local over-fitting and ill-conditioned kernel matrix. 

The results in Fig.~\ref{Fig_varyingParas_airfoil} and the Appendix indicate that VFE, SVGP and FITC provide better predictions with the increase of inducing size $m$. It is because more inducing points bring closer approximation to the full GP. This on the other hand indicates their good controllability: we could improve the predictions by simply increasing $m$.

\begin{figure}[hbt!] 
	\centering
	\includegraphics[width=0.7\textwidth]{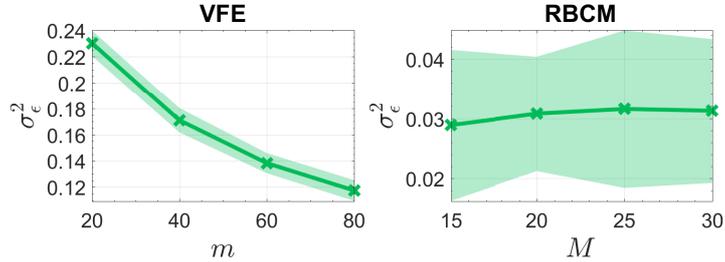}
	\caption{The estimated noise variance $\sigma^2_{\epsilon}$ of VFE and RBCM using different model parameters on the \textit{airfoil} dataset. The shaded region represents two times the standard deviation over ten runs.}
	\label{Fig_sn2_VFEvsRBCM_airfoil} 
\end{figure}

However, it is interesting to observe that the RBCM provides similar predictions with different $M$ values on not only the \textit{airfoil} dataset but also the \textit{protein} and \textit{sarcos} datasets in the Appendix. This is caused by the local attention mechanism which considers local features, thus enabling RBCM to provide better and more robust estimation of the hyperparameters in comparison to VFE and FITC. More precisely, Fig.~\ref{Fig_sn2_VFEvsRBCM_airfoil} depicts the estimated noise variance $\sigma^2_{\epsilon}$ of VFE and RBCM using different model parameters on the \textit{airfoil} dataset. As the ground truth, the noise variance estimated by the full GP on this dataset has an average value of $\sigma^2_{\epsilon}=0.0218$. It is observed that by using different $M$ values, the RBCM is always capable of providing a small $\sigma^2_{\epsilon}$ close to that of full GP. On the contrary, the VFE provides a too conservative $\sigma^2_{\epsilon}$ when using a small inducing size. The estimated $\sigma^2_{\epsilon}$ becomes closer to that of full GP with increasing $m$ at the cost of higher computing complexity.

In comparison to the conservative VFE, SVGP and FITC, the local attention helps RBCM quickly obtain a good estimation of $\sigma^2_{\epsilon}$, which in turn encourages better predictions on some datasets. However, this is often not the case for complicated datasets with heteroscedastic noise, like the \textit{chem} and \textit{sdss} datasets. In these datasets, some subregions with small noise variances favor a small estimation of $\sigma^2_{\epsilon}$, which however is not beneficial for other subregions with large noise variances. This inbalance becomes more serious when $M$ is large, since now we are forced to handle more localized regions. For example, as shown in Fig.~\ref{Fig_varyingParas_chem} in the Appendix, the RBCM provides poorer predictions with the increase of $M$.

The inconsistent performance of RBCM w.r.t. the number of experts on the five datasets indicates a poor controllability of RBCM, since it is unclear how to improve its performance by tuning $M$.

\subsubsection{Impact of the initialization of hyperparameters}
It is known that we train the GP by maximizing the marginal likelihood $p(\bm{y})$, which however is usually a non-convex optimization problem. Hence, this section investigates the robustness of scalable GPs, including VFE, FITC and RBCM, to the initialization of hyperparameters, including the SE kernel parameters $\{ l_1, \cdots, l_d, \sigma^2_f \}$ and the noise variance $\sigma^2_{\epsilon}$, on the \textit{airfoil} dataset. Particularly, we randomly initialize the kernel parameters and the noise variance as $l_i \sim \mathrm{random}(0,1)$, $\sigma^2_f \sim \mathrm{random}(0,1)$, and $\sigma^2_{\epsilon} \sim \mathrm{random}(0,0.5)$ in order to produce 100 instances.

\begin{figure}[hbt!] 
	\centering
	\includegraphics[width=0.7\textwidth]{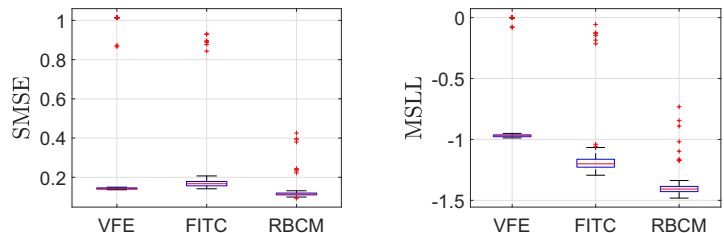}
	\caption{Impact of the initialization of hyperparameters on the performance of VFE, FITC and RBCM on the \textit{airfoil} dataset.}
	\label{Fig_airfoil_varyParas} 
\end{figure}

Fig.~\ref{Fig_airfoil_varyParas} depicts the boxplots of the three scalable GPs with respect to 100 initializations of hyperparameters on the \textit{airfoil} dataset. Some of the outliers in the boxplots represent the failure runs since they provide poor SMSE (close to one) and MSLL (close to zero). Here, we define a failure run  wherein the SMSE is larger than 0.8 and the MSLL value is larger than -0.3. It is observed that the VFE fails in 19 out of the 100 runs, the FITC fails in 7 runs, and the RBCM has no failure run. The results indicate that the RBCM is robust to various initializations of hyperparameters, because (i) as explained before, it uses the local attention to help estimate the hyperparameters well; and (ii) compared to VFE and FITC which have a large parameter space by considering the additional inducing parameters, the RBCM has a narrow parameter space due to the sharing of hyperparameters across experts.

\section{Conclusions}
\label{sec_cons}
This paper studies representative scalable GPs including the sparse approximations and the local aggregations on two toy examples and five large-scale real-world datasets. We summarize below their characteristics in terms of scalability, capability, robustness and controllability.

For sparse approximations including the prior and posterior approximations, we have the following findings from the numerical experiments:
\begin{itemize}
	\item In terms of scalability, all the sparse approximations except SVGP have the same time complexity of $\mathcal{O}(nm^2)$. The SVGP reorganizes the variational lower bound such that it factorizes over data points, thus reducing the complexity to $\mathcal{O}(bm^3)$ via SGD;
	\item In terms of capability, most of the prior approximations provide poorer predictions than the posterior counterparts. Particularly, the FITC captures heteroscedastic noise at the cost of worsening the accuracy of prediction mean. The posterior approximations including VFE and SVGP are preferred since they are faithful approximations of full GP.
	\item In terms of robustness, the sparse approximations are sensitive to the initialization of hyperparameters, because of the augmented parameter space by considering inducing and variational parameters.
	\item In terms of controllability, it is observed that VFE, SVGP and FITC generally offer better predictions with increasing $m$.
\end{itemize}

For local aggregations including GPoE and RBCM, we have the following findings from the numerical experiments:
\begin{itemize}
	\item In terms of scalability, the training complexity of local aggregations is the same as most sparse approximations when we have the training size $m_0 = m$ for each expert. But the test complexity is a bit higher since we need the predictions from $M$ experts.
	\item In terms of capability, the GPoE that allows individual hyperparameters for experts is capable of capturing non-stationary features. However, with increasing $M$, the local over-fitting and the possible ill-conditioned kernel matrix would significantly degrade GPoE's generalization capability, rendering it impractical. Contrarily, the RBCM (i) shares hyperparameters across experts to guard against over-fitting and (ii) estimates the hyperparameters well due to the local attention mechanism.
	\item In terms of robustness, the RBCM is robust to the initialization of hyperparameters on the \textit{airfoil} dataset due to the local attention mechanism.
	\item In terms of controllability, it is unclear how to tune the number $M$ of experts for RBCM.
\end{itemize}

\begin{figure}[hbt!] 
	\centering
	\includegraphics[width=0.6\textwidth]{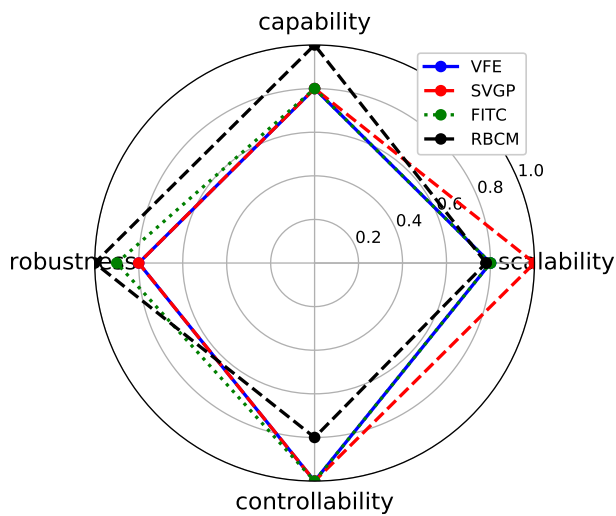}
	\caption{The radar plot to illustrate the representative global and local scalable GPs in terms of scalability, capability, robustness and controllability.}
	\label{Fig_scalable_GPs_radar} 
\end{figure}

According to the above conclusions, Fig.~\ref{Fig_scalable_GPs_radar} summarizes the characteristics of four successful global/local scalable GPs including VFE, SVGP, FITC and RBCM in terms of scalability, capability, robustness and controllability. To further improve the model capability while retaining the scalability, alternatively, we may combine sparse and local approximations together such that the hybrid could (i) guard against local over-fitting, and (ii) capture non-stationary features. Another promising avenue is the combination of scalable GPs and the well-known feature extractor, deep neural networks, to further boost the representational capability and scalability for big data~\cite{huang2015scalable, wilson2016deep}.

\section*{Acknowledgments}
This work was conducted within the Rolls-Royce@NTU Corporate Lab with support from the National Research Foundation (NRF) Singapore under the Corp Lab@University Scheme. It is also partially supported by the Data Science and Artificial Intelligence Research Center (DSAIR) and the School of Computer Science and Engineering at Nanyang Technological University.

\begin{appendices}
	\section{Appendix}
	Fig.s~\ref{Fig_varyingParas_protein}-\ref{Fig_varyingParas_sdss} below depict the impact of varying inducing size $m$ or number $M$ of experts on the performance of scalable GPs on the \textit{protein}, \textit{sarcos}, \textit{chem} and \textit{sdss} datasets, respectively.
	
	\begin{figure}[hbt!] 
		\centering
		\includegraphics[width=1.0\textwidth]{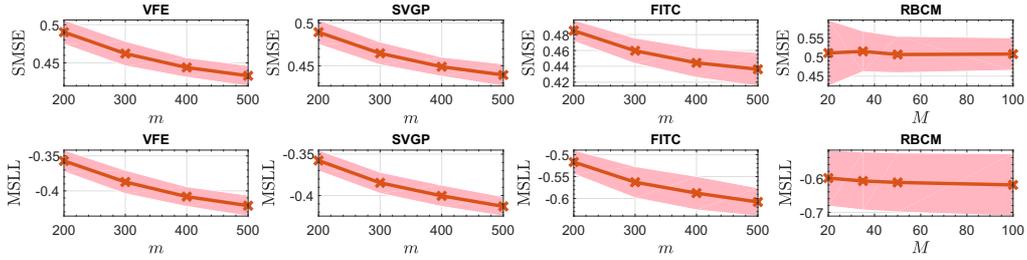}
		\caption{Impact of varying inducing size $m$ or number $M$ of experts on the performance of scalable GPs on the \textit{protein} dataset.}
		\label{Fig_varyingParas_protein} 
	\end{figure}
	
	\begin{figure}[hbt!] 
		\centering
		\includegraphics[width=1.0\textwidth]{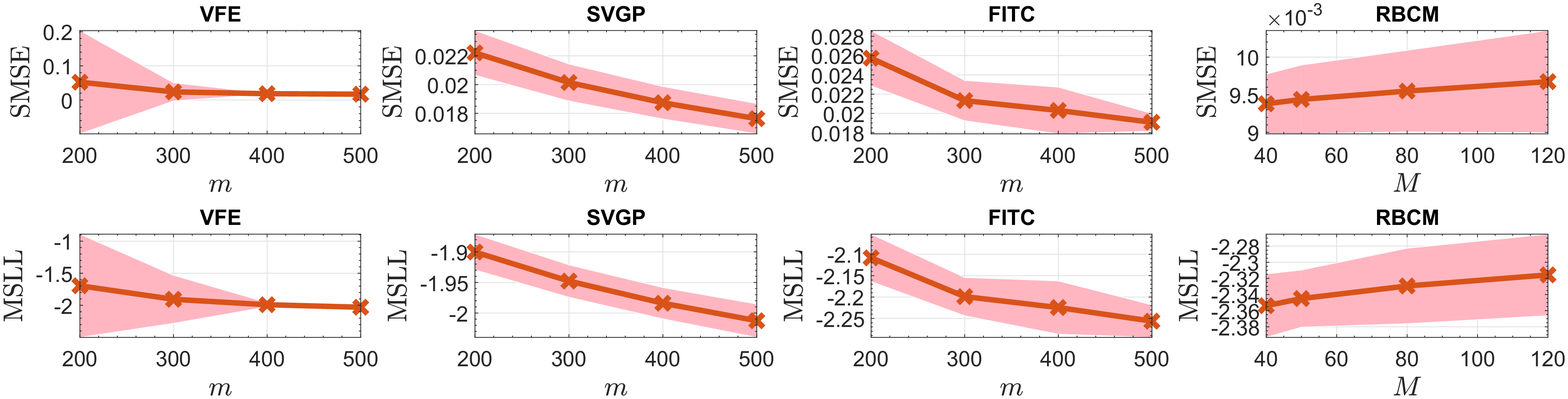}
		\caption{Impact of varying inducing size $m$ or number $M$ of experts on the performance of scalable GPs on the \textit{sarcos} dataset.}
		\label{Fig_varyingParas_sarcos} 
	\end{figure}
	
	\begin{figure}[hbt!] 
		\centering
		\includegraphics[width=1.0\textwidth]{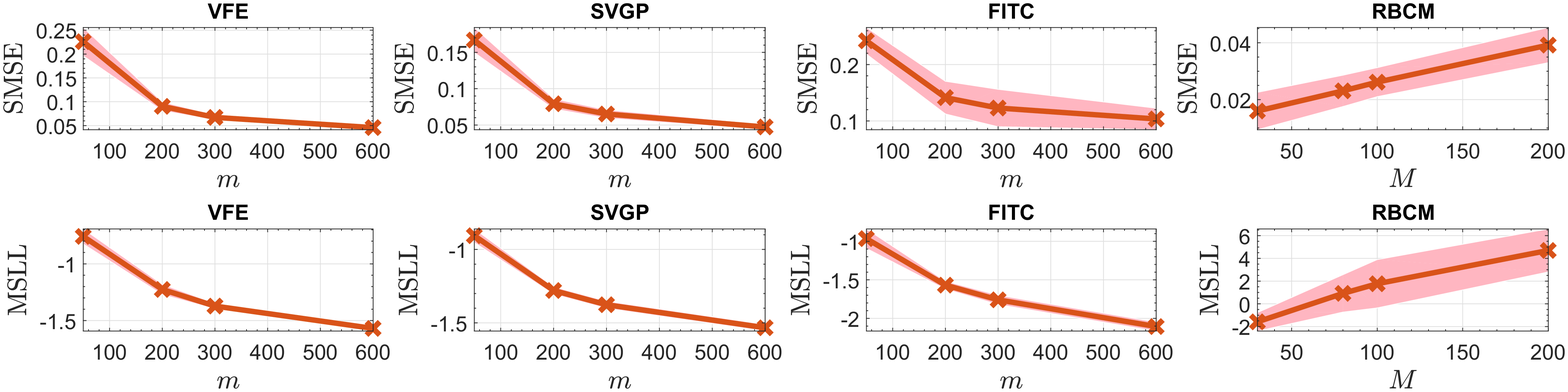}
		\caption{Impact of varying inducing size $m$ or number $M$ of experts on the performance of scalable GPs on the \textit{chem} dataset.}
		\label{Fig_varyingParas_chem} 
	\end{figure}
	
	\begin{figure}[hbt!] 
		\centering
		\includegraphics[width=1.0\textwidth]{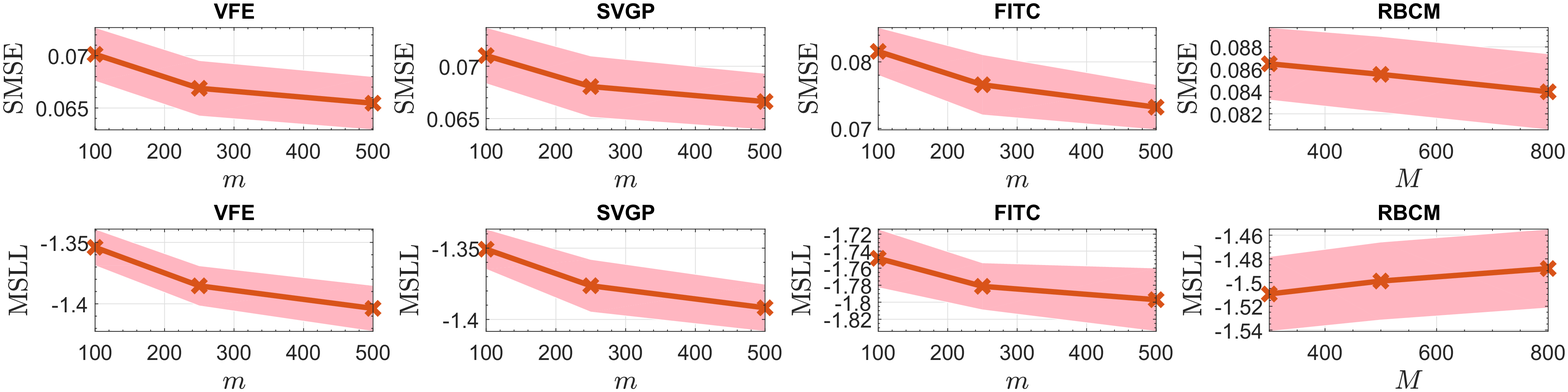}
		\caption{Impact of varying inducing size $m$ or number $M$ of experts on the performance of scalable GPs on the \textit{sdss} dataset.}
		\label{Fig_varyingParas_sdss} 
	\end{figure}
	
\end{appendices}

\section*{References}
\footnotesize
\bibliography{bigGP_comparison}

\begin{thebibliography}{10}
\expandafter\ifx\csname url\endcsname\relax
  \def\url#1{\texttt{#1}}\fi
\expandafter\ifx\csname urlprefix\endcsname\relax\def\urlprefix{URL }\fi
\expandafter\ifx\csname href\endcsname\relax
  \def\href#1#2{#2} \def\path#1{#1}\fi

\bibitem{duchaine2009computational}
F.~Duchaine, T.~Morel, L.~Gicquel, Computational-fluid-dynamics-based {K}riging
  optimization tool for aeronautical combustion chambers, AIAA Journal 47~(3)
  (2009) 631--645.

\bibitem{liu2014modeling}
X.~Liu, Q.~Zhu, H.~Lu, Modeling multiresponse surfaces for airfoil design with
  multiple-output-{G}aussian-process regression, Journal of Aircraft 51~(3)
  (2014) 740--747.

\bibitem{amrit2016efficient}
A.~Amrit, L.~Leifsson, S.~Koziel, Y.~A. Tesfahunegn, Efficient multi-objective
  aerodynamic optimization by design space dimension reduction and
  co-{K}riging, in: 17th AIAA/ISSMO Multidisciplinary Analysis and Optimization
  Conference, AIAA, 2016, pp. AIAA 2016--3515.

\bibitem{wagle2017forward}
N.~Wagle, E.~W. Frew, Forward adaptive transfer of {G}aussian process
  regression, Journal of Aerospace Information Systems 14~(4) (2017) 214--231.

\bibitem{rasmussen2006gaussian}
C.~E. Rasmussen, C.~K. Williams, Gaussian processes for machine learning, MIT
  Press, 2006.

\bibitem{liaw2002classification}
A.~Liaw, M.~Wiener, et~al., Classification and regression by random{F}orest, R
  news 2~(3) (2002) 18--22.

\bibitem{lecun2015deep}
Y.~LeCun, Y.~Bengio, G.~Hinton, Deep learning, Nature 521~(May) (2015)
  436--444.

\bibitem{neal2012bayesian}
R.~M. Neal, Bayesian learning for neural networks, Vol. 118, Springer Science
  \& Business Media, 2012.

\bibitem{liu2018survey}
H.~Liu, Y.-S. Ong, J.~Cai, A survey of adaptive sampling for global
  metamodeling in support of simulation-based complex engineering design,
  Structural and Multidisciplinary Optimization 57~(1) (2018) 393--416.

\bibitem{foreman2017fast}
D.~Foreman-Mackey, E.~Agol, S.~Ambikasaran, R.~Angus, Fast and scalable
  {G}aussian process modeling with applications to astronomical time series,
  The Astronomical Journal 154~(6) (2017) 220.

\bibitem{snoek2015scalable}
J.~Snoek, O.~Rippel, K.~Swersky, R.~Kiros, N.~Satish, N.~Sundaram, M.~Patwary,
  M.~Prabhat, R.~Adams, Scalable {B}ayesian optimization using deep neural
  networks, in: International Conference on Machine Learning, 2015, pp.
  2171--2180.

\bibitem{deisenroth2015distributed}
M.~P. Deisenroth, J.~W. Ng, Distributed {G}aussian processes, in: International
  Conference on Machine Learning, JMLR. org, 2015, pp. 1481--1490.

\bibitem{chalupka2013framework}
K.~Chalupka, C.~K. Williams, I.~Murray, A framework for evaluating
  approximation methods for {G}aussian process regression, Journal of Machine
  Learning Research 14~(Feb) (2013) 333--350.

\bibitem{quinonero2005unifying}
J.~Qui{\~n}onero-Candela, C.~E. Rasmussen, A unifying view of sparse
  approximate {G}aussian process regression, Journal of Machine Learning
  Research 6~(Dec) (2005) 1939--1959.

\bibitem{snelson2006sparse}
E.~Snelson, Z.~Ghahramani, Sparse {G}aussian processes using pseudo-inputs, in:
  Advances in Neural Information Processing Systems, 2006, pp. 1257--1264.

\bibitem{titsias2009variational}
M.~K. Titsias, Variational learning of inducing variables in sparse {G}aussian
  processes, in: Artificial Intelligence and Statistics, 2009, pp. 567--574.

\bibitem{hensman2013gaussian}
J.~Hensman, N.~Fusi, N.~D. Lawrence, Gaussian processes for big data, in:
  Uncertainty in Artificial Intelligence, AUAI Press, 2013, pp. 282--290.

\bibitem{bui2014tree}
T.~D. Bui, R.~E. Turner, Tree-structured {G}aussian process approximations, in:
  Advances in Neural Information Processing Systems, 2014, pp. 2213--2221.

\bibitem{hinton2002training}
G.~E. Hinton, Training products of experts by minimizing contrastive
  divergence, Neural Computation 14~(8) (2002) 1771--1800.

\bibitem{tresp2000bayesian}
V.~Tresp, A {B}ayesian committee machine, Neural Computation 12~(11) (2000)
  2719--2741.

\bibitem{cao2014generalized}
Y.~Cao, D.~J. Fleet, Generalized product of experts for automatic and
  principled fusion of {G}aussian process predictions, arXiv preprint
  arXiv:1410.7827.

\bibitem{liu2018generalized}
H.~Liu, J.~Cai, Y.~Ong, Y.~Wang, Generalized robust {B}ayesian committee
  machine for large-scale {G}aussian process regression, in: International
  Conference on Machine Learning, JMLR. org, 2018, pp. 1--10.

\bibitem{samo2016string}
Y.-L.~K. Samo, S.~J. Roberts, String and membrane {G}aussian processes, Journal
  of Machine Learning Research 17~(1) (2016) 4485--4571.

\bibitem{smola2001sparse}
A.~J. Smola, P.~L. Bartlett, Sparse greedy {G}aussian process regression, in:
  Advances in Neural Information Processing Systems, 2001, pp. 619--625.

\bibitem{seeger2003fast}
M.~Seeger, C.~Williams, N.~Lawrence, Fast forward selection to speed up sparse
  {G}aussian process regression, in: Artificial Intelligence and Statistics,
  PMLR, 2003, pp. EPFL--CONF--161318.

\bibitem{dezfouli2015scalable}
A.~Dezfouli, E.~V. Bonilla, Scalable inference for {G}aussian process models
  with black-box likelihoods, in: Advances in Neural Information Processing
  Systems, 2015, pp. 1414--1422.

\bibitem{wilson2015kernel}
A.~Wilson, H.~Nickisch, Kernel interpolation for scalable structured {G}aussian
  processes ({KISS-GP}), in: International Conference on Machine Learning,
  2015, pp. 1775--1784.

\bibitem{rasmussen2002infinite}
C.~E. Rasmussen, Z.~Ghahramani, Infinite mixtures of {G}aussian process
  experts, in: Advances in Neural Information Processing Systems, 2002, pp.
  881--888.

\bibitem{yuksel2012twenty}
S.~E. Yuksel, J.~N. Wilson, P.~D. Gader, Twenty years of mixture of experts,
  IEEE Transactions on Neural Networks and Learning Systems 23~(8) (2012)
  1177--1193.

\bibitem{williams2001using}
C.~K. Williams, M.~Seeger, Using the {N}ystr{\"o}m method to speed up kernel
  machines, in: Advances in Neural Information Processing Systems, 2001, pp.
  682--688.

\bibitem{silverman1985some}
B.~W. Silverman, Some aspects of the spline smoothing approach to
  non-parametric regression curve fitting, Journal of the Royal Statistical
  Society. Series B (Methodological) 47~(1) (1985) 1--52.

\bibitem{wahba1999bias}
G.~Wahba, X.~Lin, F.~Gao, D.~Xiang, R.~Klein, B.~Klein, The bias-variance
  tradeoff and the randomized {GACV}, in: Advances in Neural Information
  Processing Systems, 1999, pp. 620--626.

\bibitem{csato2002sparse}
L.~Csat{\'o}, M.~Opper, Sparse on-line {G}aussian processes, Neural Computation
  14~(3) (2002) 641--668.

\bibitem{snelson2007local}
E.~Snelson, Z.~Ghahramani, Local and global sparse {G}aussian process
  approximations, in: Artificial Intelligence and Statistics, PMLR, 2007, pp.
  524--531.

\bibitem{titsias2009model}
M.~K. Titsias, Variational model selection for sparse {G}aussian process
  regression, Tech. rep., University of Manchester (2009).

\bibitem{bauer2016understanding}
M.~Bauer, M.~van~der Wilk, C.~E. Rasmussen, Understanding probabilistic sparse
  {G}aussian process approximations, in: Advances in Neural Information
  Processing Systems, 2016, pp. 1533--1541.

\bibitem{matthews2016sparse}
A.~G. d.~G. Matthews, J.~Hensman, R.~Turner, Z.~Ghahramani, On sparse
  variational methods and the {K}ullback-{L}eibler divergence between
  stochastic processes, Journal of Machine Learning Research 51 (2016)
  231--239.

\bibitem{hoffman2013stochastic}
M.~D. Hoffman, D.~M. Blei, C.~Wang, J.~Paisley, Stochastic variational
  inference, Journal of Machine Learning Research 14~(1) (2013) 1303--1347.

\bibitem{zeiler2012adadelta}
M.~D. Zeiler, {ADADELTA}: {An} adaptive learning rate method, arXiv preprint
  arXiv:1212.5701.

\bibitem{kingma2014adam}
D.~P. Kingma, J.~Ba, Adam: {A} method for stochastic optimization, arXiv
  preprint arXiv:1412.6980.

\bibitem{hoang2015unifying}
T.~N. Hoang, Q.~M. Hoang, B.~K.~H. Low, A unifying framework of anytime sparse
  {G}aussian process regression models with stochastic variational inference
  for big data, in: International Conference on Machine Learning, 2015, pp.
  569--578.

\bibitem{chen2009bagging}
T.~Chen, J.~Ren, Bagging for {G}aussian process regression, Neurocomputing
  72~(7-9) (2009) 1605--1610.

\bibitem{okadome2013fast}
Y.~Okadome, Y.~Nakamura, Y.~Shikauchi, S.~Ishii, H.~Ishiguro, Fast
  approximation method for {G}aussian process regression using hash function
  for non-uniformly distributed data, in: International Conference on
  Artificial Neural Networks, Springer, 2013, pp. 17--25.

\bibitem{van2015optimally}
B.~van Stein, H.~Wang, W.~Kowalczyk, T.~B{\"a}ck, M.~Emmerich, Optimally
  weighted cluster {K}riging for big data regression, in: International
  Symposium on Intelligent Data Analysis, Springer, 2015, pp. 310--321.

\bibitem{van2017cluster}
B.~van Stein, H.~Wang, W.~Kowalczyk, M.~Emmerich, T.~B{\"a}ck, Cluster-based
  {K}riging approximation algorithms for complexity reduction, arXiv preprint
  arXiv:1702.01313.

\bibitem{szabo2017asymptotic}
B.~Szabo, H.~van Zanten, An asymptotic analysis of distributed nonparametric
  methods, arXiv preprint arXiv:1711.03149.

\bibitem{mair2018distributed}
S.~Mair, U.~Brefeld, Distributed robust {G}aussian process regression,
  Knowledge and Information Systems 55~(2) (2018) 415--435.

\bibitem{gal2014distributed}
Y.~Gal, M.~van~der Wilk, C.~E. Rasmussen, Distributed variational inference in
  sparse {G}aussian process regression and latent variable models, in: Advances
  in Neural Information Processing Systems, Curran Associates, Inc., 2014, pp.
  3257--3265.

\bibitem{dai2014gaussian}
Z.~Dai, A.~Damianou, J.~Hensman, N.~Lawrence, Gaussian process models with
  parallelization and {GPU} acceleration, arXiv preprint arXiv:1410.4984.

\bibitem{vanhatalo2010approximate}
J.~Vanhatalo, V.~Pietil{\"a}inen, A.~Vehtari, Approximate inference for disease
  mapping with sparse {G}aussian processes, Statistics in Medicine 29~(15)
  (2010) 1580--1607.

\bibitem{lee2017hierarchically}
B.-J. Lee, J.~Lee, K.-E. Kim, Hierarchically-partitioned {G}aussian process
  approximation, in: Artificial Intelligence and Statistics, 2017, pp.
  822--831.

\bibitem{hensman2016variational}
J.~Hensman, N.~Durrande, A.~Solin, Variational fourier features for {G}aussian
  processes, arXiv preprint arXiv:1611.06740.

\bibitem{Dua:2017}
D.~Dheeru, E.~Karra~Taniskidou, \href{http://archive.ics.uci.edu/ml}{{UCI}
  machine learning repository} (2017).
\newline\urlprefix\url{http://archive.ics.uci.edu/ml}

\bibitem{malshe2007theoretical}
M.~Malshe, L.~Raff, M.~Rockley, M.~Hagan, P.~M. Agrawal, R.~Komanduri,
  Theoretical investigation of the dissociation dynamics of vibrationally
  excited vinyl bromide on an ab initio potential-energy surface obtained using
  modified novelty sampling and feedforward neural networks. {II}. {N}umerical
  application of the method, The Journal of Chemical Physics 127~(13) (2007)
  134105.

\bibitem{almosallam2016gpz}
I.~A. Almosallam, M.~J. Jarvis, S.~J. Roberts, {GP}z: non-stationary sparse
  {G}aussian processes for heteroscedastic uncertainty estimation in
  photometric redshifts, Monthly Notices of the Royal Astronomical Society
  462~(1) (2016) 726--739.

\bibitem{huang2015scalable}
W.-b. Huang, D.~Zhao, F.~Sun, H.~Liu, E.~Y. Chang, Scalable {G}aussian process
  regression using deep neural networks., in: International Joint Conference on
  Artificial Intelligence, 2015, pp. 3576--3582.

\bibitem{wilson2016deep}
A.~G. Wilson, Z.~Hu, R.~Salakhutdinov, E.~P. Xing, Deep kernel learning, in:
  Artificial Intelligence and Statistics, 2016, pp. 370--378.

\end{thebibliography}

\end{document}